\documentclass{article}

\usepackage{arxiv}
\usepackage{array}
\usepackage{times}
\usepackage{epsfig}
\usepackage{graphicx}
\usepackage{graphics}
\usepackage{amsmath}
\usepackage{amssymb}
\usepackage{multirow}
\usepackage{setspace}
\usepackage[leftcaption]{sidecap} 
\usepackage[utf8]{inputenc} 
\usepackage[T1]{fontenc}
\usepackage{url}            % simple URL typesetting
\usepackage{booktabs}       % professional-quality tables
\usepackage{amsfonts}       % blackboard math symbols
\usepackage{nicefrac}       % compact symbols for 1/2, etc.
\usepackage{microtype} 
\usepackage{graphicx}
\usepackage[utf8]{inputenc} %unicode support
\usepackage[flushleft]{threeparttable}
\usepackage{caption}
\usepackage{environ}
\usepackage{verbatim}
\usepackage{floatrow}
\usepackage{textcomp}
\usepackage{epstopdf}
\usepackage{dblfloatfix}

\usepackage[pagebackref=true,breaklinks=true,colorlinks,bookmarks=false]{hyperref}

\author{
Mahsa Paknezhad\\
\textbf{Cuong Phuc Ngo}\\
\textbf{Amadeus Aristo Winarto}\\%\thanks{}
\textbf{Alistair Cheong}\\
\textbf{Chuen Yang Beh}\\
\textbf{Jiayang Wu}\\
Bioinformatics Institute, A*STAR,\\
30 Biopolis Street, 07-01, Matrix\\
Singapore, 138671 \\
\texttt{mahsap@bii.a-star.edu.sg}\\
\And

Hwee Kuan Lee\\
Bioinformatics Institute, A*STAR,\\
30 Biopolis Street, 07-01, Matrix\\
Singapore, 138671\\
National University of Singapore\\
Singapore, 119077\\
Singapore Eye Research Institute\\
20 College Road\\
Singapore, 169856\\
\texttt{leehk@bii.a-star.edu.sg}
}
\title{Explaining Adversarial Vulnerability with a Data Sparsity Hypothesis}

\begin{document}

%%%%%%%%% TITLE
\maketitle
%%%%%%%%% ABSTRACT

\begin{abstract}
Despite many proposed algorithms to provide robustness to deep learning (DL) models, DL models remain susceptible to adversarial attacks. We hypothesize that the adversarial vulnerability of DL models stems from two factors. The first factor is data sparsity which is that in the high dimensional input data space, there exist large regions outside the support of the data distribution. The second factor is the existence of many redundant parameters in the DL models. Owing to these factors, different models are able to come up with different decision boundaries with comparably high prediction accuracy. The appearance of the decision boundaries in the space outside the support of the data distribution does not affect the prediction accuracy of the model. However, it makes an important difference in the adversarial robustness of the model. We hypothesize that the ideal decision boundary is as far as possible from the support of the data distribution.\par
In this paper, we develop a training framework to observe if DL models are able to learn such a decision boundary spanning the space around the class distributions further from the data points themselves. Semi-supervised learning was deployed during training by leveraging unlabeled data generated in the space outside the support of the data distribution. We measured adversarial robustness of the models trained using this training framework against well-known adversarial attacks and by using robustness metrics. We found that models trained using our framework, as well as other regularization methods and adversarial training support our hypothesis of data sparsity and that  models trained with these methods learn to have decision boundaries more similar to the aforementioned ideal decision boundary. We show that the unlabeled data generated by noise in our framework is almost as effective  on adversarial robustness as unlabeled data sourced from existing datasets or generated by synthesis algorithms. The code for our training framework is available at \href{https://github.com/MahsaPaknezhad/AdversariallyRobustTraining}{https://github.com/MahsaPaknezhad/AdversariallyRobustTraining}.
\end{abstract}

\section{Introduction}
Deep learning (DL) models have achieved unprecedented accuracy for many visual recognition  \cite{rawat2017deep}, speech recognition \cite{amodei2016deep}, and natural language processing tasks \cite{young2018recent}. As DL models find their way to real-world applications, an important vulnerability in them was brought to light by the introduction of adversarial examples, generated by applying human imperceptible perturbations to the original input \cite{szegedy2013intriguing}. In a white-box setup where the adversary has access to weight parameters of the DL model, different attack algorithms such as Fast Gradient Sign Method (FGSM) \cite{goodfellow2014explaining}, Projected Gradient Descent (PGD) \cite{madry2017towards}, and DeepFool \cite{moosavi2016deepfool} already exist. Adversarial training \cite{goodfellow2014explaining} is currently one of the most effective approaches to improve robustness. Efforts to explain the adversarial vulnerability of DL models suggest that correct classification only occurs on a thin manifold and most of the high dimensional input data space consists of adversarial examples \cite{goodfellow2014explaining}. We hypothesize that two factors play important roles in adversarial vulnerability of DL models. The first factor is that a large portion of the input data space lies outside the support of the data distributions, i.e. {there exists no training data points in a large portion of the input data space. As a result, the shape of the SoftMax score surfaces learbed by DL models is not known in regions outside the support of the data distribution}. The second factor is that DL models have many redundant parameters that allow the models to learn different possible decision boundaries, especially outside the support of the input data distribution. These different decision boundaries may have similar prediction accuracy, but they may show major differences when it comes to adversarial robustness. We hypothesize that an ideal SoftMax score surface is smooth and its decision boundary will stay as far from the support of each class distribution as possible.\\
Semi-supervised learning (SSL) has been widely used to train models with higher accuracy when limited labeled data is available. SSL has also recently been deployed to train adversarially robust DL models by regularizing DL models using unlabeled datasets. These methods generate unlabeled data points using Gaussian noise and synthesis algorithms or source data points from existing datasets. The idea is that the unlabeled data provides more points in the high dimensional input data space at which the gradient of the SoftMax score surface can be regularized. In this paper, we also leverage unlabeled data, carefully generated to cover a large area around the support of the data distributions to remove the first factor in our hypothesis that is data sparsity. We then use the unlabeled data and a training framework equipped with regularization  to train models with decision boundaries similar to the described ideal decision boundary. We will analyse if models trained to have decision boundaries similar to the ideal decision boundary described above are more robust to adversarial attacks.  

Our contributions are listed as follows:
\begin{enumerate}
\item Our hypothesis about the existence of large factions in data space without training data is similar to that in the ongoing independent efforts on Open Set Recognition (OSR) \cite{scheirer2012toward}, whose objective is to train models that can identify data points from seen classes in the training data or Known Known Classes (KKCs) accurately, while rejecting data points from unseen classes or Unknown Unknown Classes (UUCs) that are passed into the model during testing. In these algorithms, it is assumed that UUCs are located in a subspace far from KKCs in the high dimensional input data space and the model is trained to have decision boundaries that bind the support of each KKC distribution tightly in order to reserve the space outside the support of the data distribution for UUCs.
\item We hypothesize that adversarial vulnerability of DL models for high dimensional input data stems from two factors. The first factor is data sparsity in the high dimensional input data space. The second factor is that DL models with a large number of parameters tend to learn many different SoftMax score surfaces with decision boundaries that get close to the class distributions in the input data space.  We develop a training framework for DL models to learn SoftMax score surfaces similar to what we believe is the ideal SoftMax score surface for DL models. We define the ideal SoftMax score surface as the SoftMax score surface that is smooth and its decision boundary lies far from the support of each class distribution. The training framework generates unlabeled data in a novel way to alleviate data sparsity in a large area around the labeled dataset in the high dimensional input data space. Using the generated unlabeled data and a regularization term, the training framework encourages DL models to learn SoftMax score surfaces similar to the ideal SoftMax score surface. We show that the relative distance of the decision boundary for DL models trained with our training framework using Jacobian regularizer is larger compared to DL models trained with the Jacobian regularizer alone for MNIST, CIFAR10 and Imagenette datasets.
\item We show that our training framework is able to train DL models with SoftMax score surfaces similar to the ideal SoftMax score surface for a 2D points dataset. 
\item We show that unlabeled data generated by Gaussian noise generated by our training framework is almost as effective as unlabeled data generated by more costly and complex synthesis algorithms or those sourced from other existing datasets.
\item We compare adversarial robustness of DL models trained using Gaussian-generated unlabeled data and the Jacobian regularizer with intermediate noise injection and the Jacobian regularizer and show that injecting noise at the intermediate layers of the network following the same strategy proposed in our training framework improves adversarial robustness of models.
\item In an OSR setting, we show that models trained with our training framework are consistently less confident in predicting data points from unknown classes which is the desirable behaviour.
\end{enumerate}
There are a number of works that utilize unlabeled or labeled data generated by Gaussian noise for training adversarially robust models. These include the work by Zhang et al. \cite{zhang2019defense} where instead of generating adversarial examples using the direction from the data point to the decision boundary, perturbations are generated for adversarial training using inter-sample relationships in the input data batch. Also, in the works by Chernyak et al. \cite{chernyak2021constant}, Zantedeschi et al. \cite{zantedeschi2017efficient} and Li et al. \cite{li2020toward} models trained with new data points drawn from a neighborhood around each training example with the same label as the original data point are found to be more robust to adversarial attacks. None of these methods ensure that the generated data distribution covers a large area around the support of the labeled data distribution. However, they do support our hypothesis that data sparsity is one of the factors that makes DL models prone to adversarial attacks. \par
In summary, in this paper we first propose two hypotheses: 1) there exists data sparsity in high dimensional input data space. As a result, the shape of the SoftMax score surface learned by DL models is unknown in the area outside the support of the input data distribution and 2) the large number of parameters in DL models allows DL models to learn many possible SoftMax score surfaces in the area outside the support of the input data distribution. Our method is related to our hypothesis by generating unlabeled data that alleviates data sparsity in the high-dimensional input data space and training DL models to learn a unique SoftMax score surface that we believe is an ideal SoftMax score surface for DL models. Training DL models to learn the ideal SoftMax score surface was performed using the generated unlabeled data and a regularization term. \par

We show that a model trained with the generated unlabeled data and a regularization term can learn a SoftMax score surface similar to the ideal SoftMax score surface for a 2D points dataset. We show that the relative distance of the decision boundary of models trained with our training framework increased compared to normal training and is on par with the relative distance of the decision boundary of models trained with other regularization methods. We also show that the prediction confidence of models on UUC data points is consistently lower for models trained with our training framework compared to models trained without regularization implying that the SoftMax score surface at the location of UUC distributions in the high dimensional input data space is flattened. These results imply that the SoftMax score surface of the trained models with our training framework is similar to the ideal SoftMax score surface.

\section{Related Work}
\label{sec:relatedwork}

Over the last few years, many algorithms have been proposed to improve adversarial robustness of DL models. A number of such algorithms propose to detect adversarial attacks. The work by Goswami et al. \cite{goswami2018unravelling} is an example of such algorithms in which attacks are detected by observing the response of hidden layers in the network or the work by Grosse et al. \cite{grosse2017statistical} where adversarial examples are detected by taking into account that they are drawn from a different distribution as the distribution of the original data. There are many other proposed algorithms that detect adversarial examples such as the works by Peck et al. \cite{peck2020detecting}, Gong et al. \cite{gong2017adversarial}, Feinman et al. \cite{feinman2017detecting}, Das et al. \cite{das2017keeping}, Liang et al. \cite{liang2018detecting} and Lu et al. \cite{lu2017safetynet}. Another group of methods aim to cancel the effect of adversarial perturbations on the model's output. The work by Xie et al. \cite{xie2017mitigating} is a good example of such methods where the authors perform low-level image transformations such as random resizing and random padding on the input image before passing the image to the classification model to destroy the structure of adversarial perturbations. The work by Liao et al. \cite{liao2018defense} proposes a denoiser based on the observation that while the difference between the clean and adversarial sample is small, this difference is large in the high-level representation of the two samples in the classifier. The proposed denoiser is trained to suppress the influence of adversarial perturbations by minimizing the distance between the high-level representation of the two images. In another work, Xie et al. \cite{xie2019feature} propose a feature denoising network that when combined with adversarial training significantly improves adversarial robustness of the trained model. Methods that are built upon adversarial training, such as the work by Baytas et al. \cite{baytas2021robustness} are another group of algorithms. In the work by Baytas et al. \cite{baytas2021robustness} adversarial training is done using perturbations that are automatically generated by a generator network or in the work by Tram{\`e}r et al. \cite{tramer2017ensemble} perturbations are transferred from other models. Qin etl al. \cite{qin2019adversarial} perform adversarial training while treating correctly-classified and misclassified examples differently. Apart from these methods, there are other approaches to train robust models such as the work by Theagarajan et al. \cite{theagarajan2019shieldnets} where adversarial-free zones around the input data distribution of the model are found and adversarial examples are projected to the adversarial-free zones. In the work by Wong et al. \cite{wong2018provable}, a convex outer bound is constructed by applying norm-bounded perturbations on the training data. The authors compute and optimize over the worst case loss within this convex outer bound to train a model that is provably robust to any norm-bounded adversarial attacks. Being computationally heavy, the work by Wong et al. \cite{wong2018provable} was further modified to scale to larger networks \cite{wong2018scaling,wang2018mixtrain}. Dapello et al.  \cite{dapello2020simulating} increase adversarial robustness of DL models by replacing the first few layers of the network with neural layers with fixed weights that simulate primary visual cortex. Gittings et al. \cite{gittings2020vax} propose a defence mechanism against adversarial patch attacks which leverages GANs to come up with effective adversarial patches and fine tunes the model to build resilience against adversarial patch attacks. Many other algorithms have been proposed to address adversarial vulnerability of DL models such as the works by Shaham et al. \cite{shaham2018understanding}, Li et al. \cite{li2018certified}, Yang et al. \cite{yang2019me}, Guo et al. \cite{guo2018sparse}, Pang et al. \cite{pang2019improving}, Zhou et al. \cite{zhou2019latent}, Cohen et al. \cite{cohen2019certified}, Goel et al. \cite{goel2020dndnet}, Rusak et al. \cite{rusak2020simple}, and Liu and Sun \cite{liu2021alpha}.

A large number of proposed methods for training adversarially robust models are categorized as regularization methods where an additional loss term is specified for model training. Such regularization terms in the literature are designed to achieve different objectives. One objective is to smooth the decision surface of models such as the work by Simon-Gabriel et al. \cite{simon2019first}, Xu et al. \cite{xu2020adversarial} and Roth et al. \cite{roth2018adversarially}. Another objective is to keep the decision boundary away from the training data points such as the work by Singla and Feizi \cite{singla2020second}, Glimer et al. \cite{gilmer2018adversarial} and Zhang et al. \cite{zhang2019theoretically}. In the following, we will provide a brief review of related literature on regularization approaches as well as SSL algorithms.
\subsection{Regularization Methods}
Existing algorithms train DL models with decision boundaries that stay far from training data points, have smooth gradients or small curvatures\cite{moosavi2019robustness}. Yan et al. \cite{yan2018deep} incorporated a DeepFool attack module in training to maximize the distance between the decision boundary of the trained model and the training data points. They added the regularization term $\lambda \sum_k R(-{||\Delta_{x_k}||_p}/{||x_k||_p})$ with a weight parameter $\lambda$ to the natural error in the loss function and jointly optimized the model over the loss function during training. The regularization term is a scale of $\Delta_{x_k}$ which is the minimum adversarial perturbation found by DeepFool for an input image $x_k$. $R$ is a monotonically increasing function. Zhang et al. \cite{zhang2019theoretically} also proposed a regularization term to increase the distance to the decision boundary by adding the boundary error $\max_{x'_k \in \mathbf{B}(x_k,\epsilon)} \phi\left(f_\theta(x_k)f_\theta(x'_k)/\lambda\right)$ as a regularization term. $B(x,\epsilon)$ represents the neighborhood of input $x_k$, and $\phi$ is a surrogate of 0-1 loss for binary classification. The regularization term minimizes the difference between the prediction for the natural image $f_\theta(x_k)$ and the adversarial example $f_\theta(x'_k)$. Jakubovitz and Giryes \cite{Jakubovitz2018improving} added the regularization term $\lambda\sqrt{\sum_{i=1}^d \sum_{j=1}^n\sum_{k=1}^N \left(\frac{\partial}{\partial x_i} z^{(L)}_j (x_k)\right)^2}$  which is the Euclidean norm of the network's Jacobian matrix evaluated on input $x_k$  where $f_\theta(x_k)=\text{SoftMax}\{z^{(L)}(x_k)\}$. The proposed regularization is applied during extended training of the model after normal training where $d$ is the dimension of the input $x_k$, $n$ is the dimension of the output of $f_\theta(x_k)$, and $N$ is the number of data points in the dataset. The reason for choosing a Jacobian matrix is said to be that Jacobian is related to the curvature of the model's decision boundary and the distance to the closest adversarial example. Moosavi et al. \cite{moosavi2019robustness} provide theoretical evidence that a small curvature in the loss function increases the robustness of the model. They propose a regularization term that reduces the curvature of the loss function by minimizing eigen values for the Hessian matrix with respect to the inputs: $\lambda \frac{1}{h^2} \mathbf{E} ||\nabla l(x_k, hz) - \nabla l(x_k)||^2 $. Here, $h$ is the discretization step over $z \sim N(0, Id)$.  Ross and Doshi-velez \cite{ross2018improving} use adversarial training and distillation \cite{ba2014do} as a regularization mechanism and use the regularization term $\lambda \sum_{i=1}^d \sum_{k=1}^N \left(\frac{\partial}{\partial x_i} \sum_{j=1}^n -y_{kj} \log f_\theta (x_k)_j\right)^2$ which encourages that the KL divergence between the predictions and the labels does not change significantly if the input changes slightly \cite{drucker1992improving}. Carmon et al. \cite{carmon2019unlabeled} use the same regularization term proposed by Zhang et al. \cite{zhang2019theoretically} except that they deploy unlabeled samples in addition to labeled samples during training.

\subsection{Semi-Supervised Learning}
\label{sec:ssl}
SSL algorithms have also been used for consistency training which refers to all the algorithms that regularize model predictions to be resilient against noise injections into input examples or hidden states. Xie et al.~\cite{xie2019unsupervised} substitute noise injection methods with advanced data augmentation methods to generate unlabeled data and use it together with labeled data for consistency training. Verma et al. \cite{verma2019manifold} proposes training the network with new minibatches generated by randomly mixing two minibatches in an intermediate layer of the network and giving the new minibatch a soft label.  Miyato et al. \cite{miyato2018virtual} extend adversarial training~\cite{goodfellow2014explaining} to use both labeled and unlabeled data points for regularizing the conditional label distribution around each input data point. Zhai et al. \cite{zhai2019adversarially} utilize SSL for adversarially robust generalization by using portions of the MNIST and CIFAR10 dataset as unlabeled data. The authors train DL models to be accurate on labeled data and robust on both labeled and unlabeled data. Uesato et al. \cite{uesato2019are} utilize the 80 Million Tiny Images dataset as unlabeled data for the labeled CIFAR10 dataset, which led to a 4\% improvement in the robust accuracy of the trained model. Robust accuracy is defined as the accuracy of the model in predicting the correct classes for adversarially perturbed images. The authors also utilized portions of CIFAR10 and SVHN as unlabeled data points for these datasets to improve the adversarial robustness of trained models. They showed that models trained with smaller labeled datasets, with most labels for samples in the CIFAR10 and SVHN dataset removed are more adversarially robust. Najafi et al. \cite{najafi2019robustness} removed labels from portions of their training dataset, assigned soft labels to the unlabeled data according to an adversarial loss, and used these soft-labeled images with the labeled images for training. Carmon et al.\cite{carmon2019unlabeled} also augmented CIFAR10 with 500K unlabeled images from the 80 Million Tiny Images dataset. On SVHN, they used the dataset's own extra training set as unlabeled images. In these studies except for noise injection methods, unlabeled data was captured in a similar expensive manner as labeled data.

\begin{figure}[t]
\centering
\includegraphics[clip=true, trim=1cm 2cm 0cm 1cm,width=\textwidth]{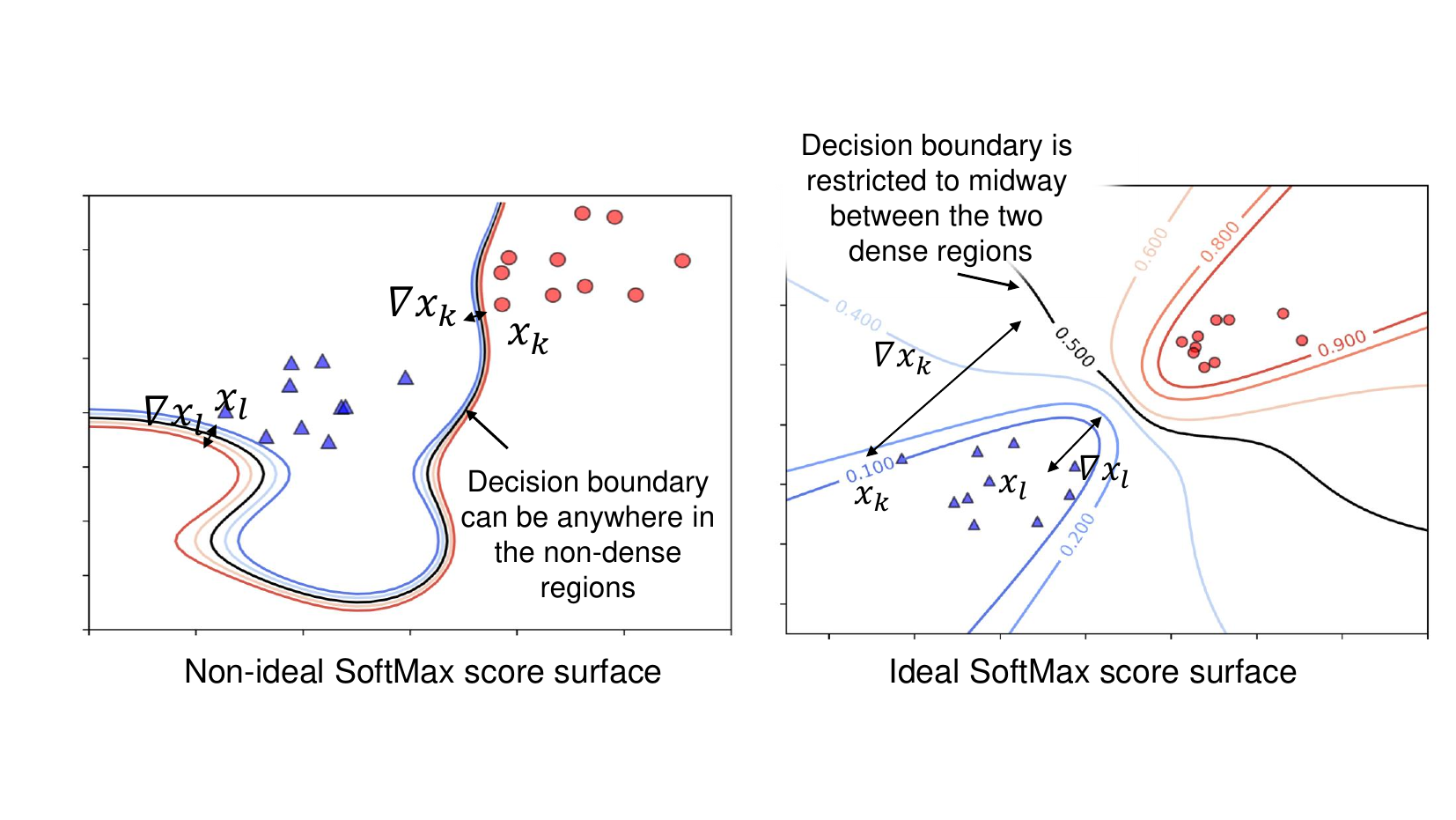}
\caption{(Left) shows a SoftMax score surface that a DL model can learn for classifying a 2D point dataset. A model with such a SoftMax score surface is very susceptible to adversarial attacks since the decision boundary gets close to the data distribution and small perturbations such as $\nabla x_k$ and $\nabla x_l$ in data points (e.g. $x_k$ and $x_l$) can make the data points cross the decision boundary. (Right) shows what we hypothesize to be the ideal SoftMax score surface which can improve the adversarial robustness of a model.  In contrast to the left figure, the decision boundary for this surface is smooth and stays far away from both class distributions. The reader is referred to the online version of this paper for the color representation of this figure.}
\label{fig:norm_ideal}
\end{figure}

\section{Method}
This section presents our developed framework for training DL models to learn decision boundaries similar to our hypothesized ideal decision boundary. We will first present the overall objective. Later, we introduce the notations used in the paper. Finally, the data generation and model training steps are described.  

\subsection{The Ideal Decision Boundary}
\label{sec:idea}
Fig.~\ref{fig:norm_ideal} shows a binary classification problem for a 2D points dataset. Also shown is the schematic representation of the SoftMax score surface for two trained DL models on the 2D points dataset using contour plots.
Fig. \ref{fig:norm_ideal} (Left) shows the SoftMax values for a DL model with a SoftMax score surface that is hypothesized to be non-ideal. Due to the short distance of its decision boundary to certain data points, an attack model will be able to find small perturbations such as $\nabla x_l$ and $\nabla x_k$ for data points (e.g. $x_k$ and $x_l$, respectively) that will make the perturbed data points cross the decision boundary and consequently be assigned to the other class label.\par
In contrast, the SoftMax score surface shown in Fig. \ref{fig:norm_ideal} (Right) is hypothesized to be more robust against adversarial attacks. The SoftMax score surface spans outwards in the space around the data distribution and gradually changes from zero to one from one class to the other. This is desirable as it keeps the decision boundary as far as possible from both class distributions, making it more difficult for adversarial attacks to fool the model via small perturbations.\par

In Fig. \ref{fig:norm_ideal} (Left), the SoftMax score surface learned by the model is very steep around the decision boundary and the decision boundary gets close to the data points. The hypothesized ideal SoftMax score surface (Fig.~\ref{fig:norm_ideal}) has two distinct characteristics: 
\begin{enumerate}
\item The first characteristic is the smooth gradient of the SoftMax score surface. The advantage of SoftMax score surfaces with smooth gradients has been studied and is well-known~\cite{Jakubovitz2018improving,ross2018improving}.
\item The other characteristic is spreading of the SoftMax scores in the space away from the class distributions. This characteristic is very beneficial as will be explained in the following paragraphs. 
\end{enumerate}
A SoftMax score surface that spreads outward in the space around the data distribution results in a decision boundary that stays far from all class distributions. Consequently, it will be more difficult for the attack model to find a small perturbation that changes the predicted label for the clean data point. \par
Another advantage of having such a SoftMax score surface is in the behaviour of the model when it receives a data point that does not belong to any of the classes the model was trained on or unknown unknown classes (UUCs). 
As an example, consider the model with the non-ideal SoftMax score surface in Fig.~\ref{fig:norm_ideal} (Left). If during testing the model receives a data point from a new class that it has not been trained on, the distribution from which that data point is sampled from may be far from the data distribution the model is trained on. If the data distribution for the unseen class is at the right side of the decision boundary in Fig.~\ref{fig:norm_ideal} (Left), the model will label the new data point with the label of the class distribution on the right with a very high confidence (e.g $>$ 90\%), whereas the proposed ideal SoftMax score surface in Fig.~\ref{fig:norm_ideal} (Right) assigns label of the class distribution on the right with a lower confidence (e.g. 60\%).\par
To train a DL model to learn such an ideal SoftMax score surface with an ideal decision boundary we sample unlabeled data points in a large neighborhood of the labeled data distribution so that the generated unlabeled dataset covers a large faction of the high dimensional input data space outside the support of the data distribution and between class distributions. We use a regularization term to to encourage small gradients on the SoftMax score surface of the DL model at the location of data points in the labeled and unlabeled datasets in the high dimensiontal input data space. Our experiments show that the learned SoftMax score surface by DL models using our training framework looks similar to what we define as the ideal SoftMax score surface for DL models.

\subsection{Notations}
Consider a $c$-class classification problem for learning a predictor $f_\theta$ with parameters $\theta$ to map inputs $x_i \in X \subset \mathbb{R}^n$ to labels $y_i \in Y = \{1,\cdots c\}$ where $n$ is the dimension of the input data space. In this case, $f_\theta$ will be of the form: $f_\theta(x) = \arg\max_{y \in Y} p_\theta(y|x)$. The labeled dataset is specified by $D = \{\left(x_i, y_i\right)| x_i \in X , y_i \in Y, i = 1, \cdots, N_l\}$, $y_i$ is the ground-truth label for $x_i$, and $N_l$ is the number of labeled data points.
The generated unlabeled dataset is specified by $U = \{u_{i,j} \in \mathbb{R}^n |i = 1, \cdots, N_l, j = 1, \cdots, N_u\}$ where $u_{i,j}$ is the $j^{th}$ unlabeled data point generated using the $i^{th}$ labeled data point, and $N_u$ is the number of generated unlabeled data points using one labeled data point. No ground-truth label is known for the unlabeled data points. The unlabeled data points are sampled from the same data space as the labeled data points, hence $u_{i,j} \in \mathbb{R}^n$. We concatenate the sets $X$ and $U$ and define $\Psi =\{\psi_k| \psi_k \in X \cup U, k=1, \cdots, N_l+N_l\times N_u\}$. Hence, a new dataset is generated:
\begin{equation}
    D_{new} = {\begin{cases}
    (\psi_k, y_k) & \text{if } \psi_k \in X\\
    \psi_k & \text{if } \psi_k \in U.
    \end{cases}
    }
    \label{eq:d_new}
\end{equation}
In the next section, we explain how unlabeled data is generated. 

\subsection{Generating Unlabeled Data Points}
Utilizing unlabeled data points to train adversarially robust neural networks is not a new idea as explained in the Section \ref{sec:ssl}. In many domains such as cancer research, obtaining even an uncurated dataset of images can be expensive and time consuming. Preparation and scanning of a single sample of cancer tissue can cost hundreds to thousands of dollars.  Also, there is no guarantee that the acquired dataset will cover a large portion of the data space around the support of the labeled data distribution.\par
In our training framework, we generate unlabeled data points very cheaply from the labeled dataset using simple additive Gaussian noise. Unlabeled data points were generated from the labeled data points by adding noise to the labeled data points sampled from an isotropic Gaussian distribution. The novelty in our training framework is in sampling Gaussian noise from a suitable Gaussian distribution that allows the generated unlabeled dataset to cover a large neighborhood around the labeled data distribution. The suitable Gaussian distribution was found by measuring the spread of the labeled data distribution in the input data space which was measured as the mean of the Euclidean distance between pairs of data points ($\mu_{pair}$) in the labeled dataset. In other words, $\mu_{pair}$ is a parameter that we measure only once from the input labeled dataset. From the input labeled dataset $X$ with $N_l$  labeled data points, pairs of data points are randomly selected ($x_i,x_j \sim X$) and the Euclidean distance between each pair is measured. The mean value of these measurements is defined as $\mu_{pair}$. We then use multiples of $\mu_{pair}$ as the standard deviation for the Gaussian distribution from which noise is sampled to generate unlabeled data from each labeled data point by adding the sampled noise to the labeled data point. Generating unlabeled data in this way increases the likelihood that the unlabeled data covers a large area around the original labeled data distribution. The standard deviation of the Gaussian distribution was specified as k-times the mean value.  The Gaussian distribution was uniform in all input data dimensions. More precisely, the co-variance matrix could be represented by the simplified matrix $\sigma^2 I$ with $\sigma$ equal to $k \times \mu_{pair}$. For each labeled data point $x_i$, a series of $N_{u}$ unlabeled data points $\{u_{i,j}|j= 1, \cdots,N_{u}\}$ were sampled from the area around the labeled data point in $\mathbb{R}^n$.  We set $k=3$ in our experiments in the paper. The effect of the value of $k$ on the robustness of the DL models is analysed in Appendix E of the Supplementary Materials. Unlabeled data points were generated once only for the labeled dataset. \par

Thus, the unlabeled dataset is generated to cover a large data space outside the support of the data distribution. We believe our generated unlabeled data will support not only the adversarial attack directions but many other directions in the input data space. Using this unlabeled dataset together with the labeled dataset, the DL model is trained to have a smooth SoftMax score surface using regularization. Using the generated unlabeled data and a training framework equipped with regularization, we experimentally show that models learn a decision boundary that is away from the support of each class distribution. We will show in Section \ref{sec:unlabeled} that using Gaussian noise to generate unlabeled data instead of collecting more data or using synthesis algorithms is almost as useful to achieve a smooth SoftMax score surface with a decision boundary away from the class distributions similar to the ideal decision boundary explained before. In the next section, we explain what regularization technique was deployed to train the model to have a smooth SoftMax score surface.\par 

\subsection{Training the CNN}
\label{sec:training}
A strong regularization term for training a model with a smooth SoftMax score surface is the Euclidean norm of the network's Jacobian matrix evaluated on each input data point as proposed by Jakubovitz et al. \cite{Jakubovitz2018improving} and introduced in Section \ref{sec:relatedwork}. We believe, regularization algorithms that impose a boundary condition such as TRADES \cite{zhang2019theoretically} may not be as effective in training models with the aforementioned ideal decision boundary as these algorithms push the decision boundary away only in the $\epsilon$-neighborhood of labeled data points. Minimizing the Euclidean norm of the Jacobian matrix for each input data point will make the SoftMax score surface smooth in each input dimension at that data point. Applying this regularization term not only on the labeled data points but also on the unlabeled data points will smoothen the SoftMax score surface even outside the support of the data distribution which aligns well with our objective. We believe this regularization term and the generated unlabeled data points which cover a large fraction of the high dimensional input data space around the original data distribution will train models similar to the aforementioned ideal decision boundary.\\
The disadvantage of this regularization term is that training models using this regularization term is very time-consuming. According to the mean value theorem, there exists two input data points $x_1, x_2 \in X$ and $x^\prime = tx_1 + (1-t)x_2, 0 \leq t \leq 1$ for which:
\begin{equation}
    \frac{||f_\theta(x_1) - f_\theta(x_2)||_2}{||x_1 - x_2||_2} \leq \sum_{k=1}^{c}||\nabla_x f_{\theta,k}(x^\prime)||_2.
    \label{eq:meanvalue}
\end{equation}
where $||.||_2$ is the Euclidean norm function. The Jacobian regularization minimizes the term at the right side of Eq. \ref{eq:meanvalue}. To decrease the training runtime in order to test our hypothesis rigorously, we implemented a more naive version of this regularization term by instead minimizing the term at the left side of the Eq. \ref{eq:meanvalue} which is the slope of the SoftMax score surface between two input data points.\par
\begin{figure}[t]
\centering
    \includegraphics[clip=true, trim=0cm 0cm 0cm 0cm, width=\textwidth]{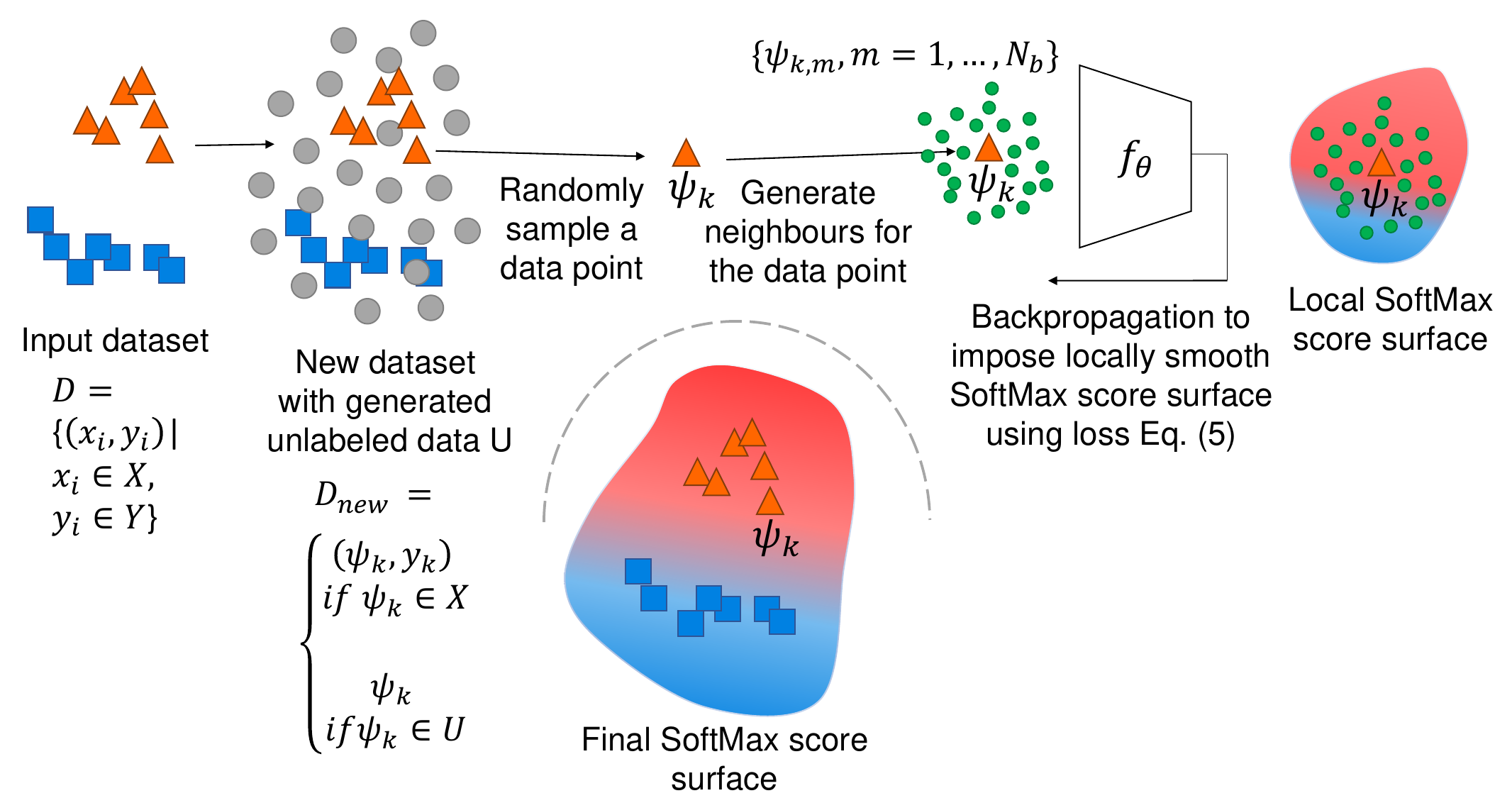}
    \caption{Figure shows the pipeline for training a DL model for the input dataset shown on the left. For the training dataset an unlabeled dataset is generated. During training a data point is sampled from either the labeled or unlabeled dataset. The local SoftMax score surface around the sampled data point and the generated neighbors for that data point is made smooth using the loss function in Eq. \ref{eq:model_loss}. The final ideal SoftMax score surface of the model for the input dataset is shown at the bottom of the figure.}
    \label{fig:pipeline}
\end{figure}

\begin{figure}[t]
\centering
    \includegraphics[clip=true, trim=0cm 12cm 6cm 0cm, width=\textwidth]{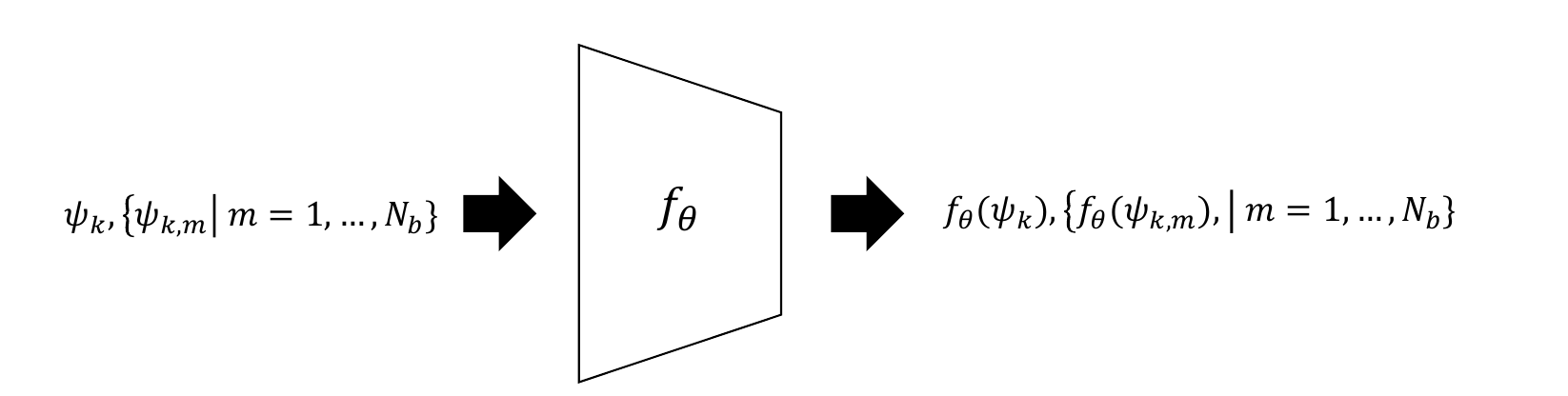}
    \caption{The input training batch to the CNN model. In each batch, a data point, $\psi_k$, together with $N_{b}$ generated neighbor data points, $\{\psi_{k,m}|m=1,\cdots, N_{b}\}$, are fed to the model. The model is trained by minimising the loss function \ref{eq:model_loss}.}
    \label{fig:cnn}
\end{figure}

 
Fig. \ref{fig:pipeline} shows a schematic representation of the training framework for a DL  model for the input dataset shown on the left. The final SoftMax score surface of the trained model using this training framework is also shown at the bottom of the figure. During the training process, a data point, $\psi_k$,  was selected randomly from the new dataset $D_{new}$ (Eq. \ref{eq:d_new}). Similar to the way unlabeled data points were generated, a specific number of neighbor data points, $N_{b}$ (similar to $N_u$ for unlabeled data points), were generated for the selected data point by adding $N_b$ noises sampled from an isotropic Gaussian distribution with 10-times smaller standard deviation than the standard deviation for generating unlabeled data. Neighbor data points were generated on the fly for regularization purposes. Hence, they were not part of the new dataset. These neighbor data points are denoted by $\{\psi_{k,m}| m=1,\cdots, N_{b}\}$. The selected data point was then fed into the model together with the generated neighbors, as shown in Fig.~\ref{fig:cnn} as a mini-batch.\par

For the selected data point $\psi_k$ and a single neighbor data point $\psi_{k,m}$, which was generated by adding small noise to $\psi_k$, the model with a smooth SoftMax score surface would ideally output similar predictions, denoted by $f_\theta(\psi_{k})$ and $f_\theta(\psi_{k,m})$, respectively. This requirement was enforced by smoothing the slope of the SoftMax score surface between the data point $\psi_k$ and each of its neighbor points $\psi_{k,m}$. This is our naive regularization term, $Loss_{reg}$, and is defined as:
\begin{align} \nonumber
    Loss_{reg}(\psi_k,\theta) = 
    \sum_{m=1}^{N_{b}} \frac{||f_\theta(\psi_{k,m}) - f_\theta(\psi_k)||_2}{||\psi_{k,m} - {\psi_k}||_2}, \\
    \forall\psi_k \in X \cup U, \{\psi_{k,m}| m=1,\cdots, N_{b}\}
\end{align}

To ensure that the model learns to correctly classify the labeled data points $\{\psi_k \in X\}$, a supervised loss function, $Loss_{ce}$, in our case cross entropy shown as $H$, was considered to compare the predicted label, $f_\theta(\psi_k)$ for the labeled data point with the ground-truth label $y_k$.
\begin{equation}
    Loss_{ce}(\psi_k, y_k, \theta) = H(y_k, f_\theta(\psi_k)), \forall \psi_k \in X
\end{equation}

The overall loss function for the CNN model is a weighted combination of the supervised and the regularization loss functions, defined as below. The cross entropy loss is measured only for the data points that came from the labeled dataset $X$ and is zero otherwise.
\begin{align}  \label{eq:model_loss}
    &Loss  = \sum_{\psi_k \in X} Loss_{ce}(\psi_k, y_k, \theta)~+ 
     \lambda \times \sum_{\psi_k \in X \cup U}  Loss_{reg}(\psi_k,\theta),
\end{align}
where $\lambda$ is a hyperparameter. We use this loss function for training DL models in the coming experiments to test our hypotheses. More details on the size of the isometric Gaussian distributions that are used to sample noise and generate unlabeled data points and neighbor data points are provided in Appendix B of the Supplementary Materials. In the next section,  we describe our experiment design. We assess our hypothesis using the proposed training framework and show the conclusions that were drawn from these experiments. \par

\begin{figure}[t]
\centering
\includegraphics[clip=true, trim=1cm 7.5cm 4.5cm 2cm,width=\textwidth]{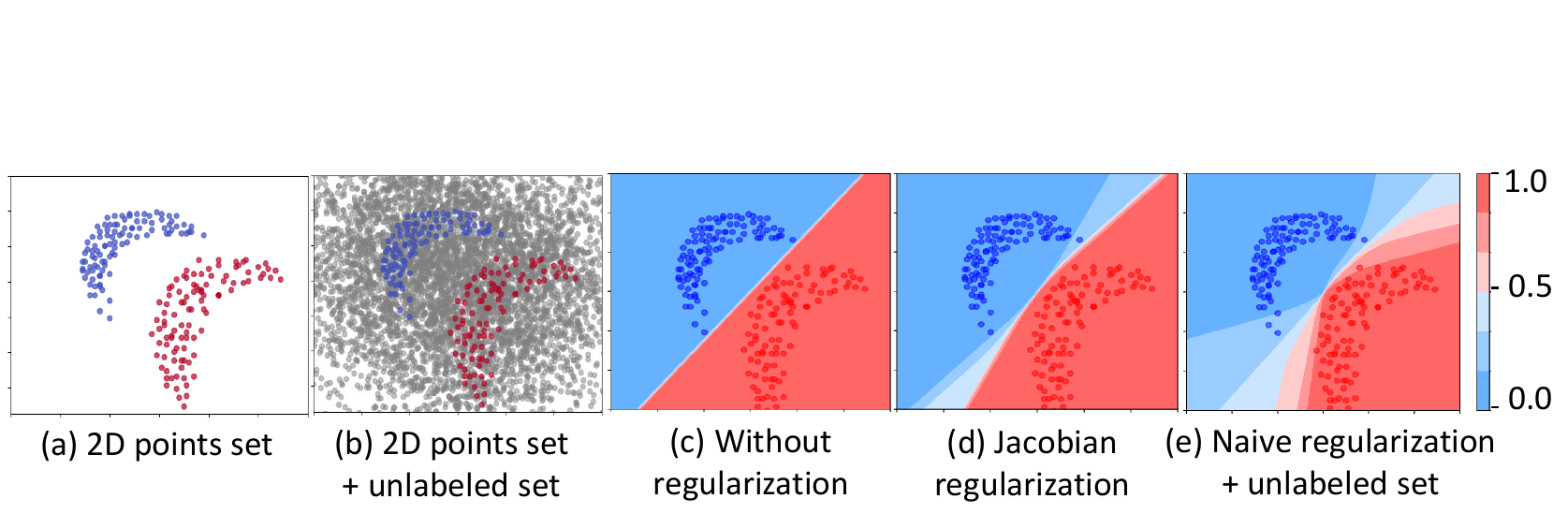}
\caption{(a) our 2D points dataset ($\{X,Y\}$). (b) generated unlabeled dataset for the labeled dataset is shown in gray ($U$). (c) the SoftMax score surface learned by a model without regularization and unlabeled data, (d) the SoftMax score surface learned by the Jacobian Regularization term \cite{Jakubovitz2018improving} without unlabeled data, (e) the SoftMax score surface learned by our naive regularization term using the unlabeled dataset. The reader is referred to the online version of this paper for the color representation of this figure.}
\label{fig:2dpoint}
\end{figure}

\section{Experiments}
\label{sec:experiments}
To test our hypothesis about the proposed ideal decision boundary, different datasets comprising of both 2D points and image datasets including MNIST \cite{lecun1998gradient}, CIFAR10 \cite{krizhevsky2009learning}, and Imagenette which is a subset of the well-known ImageNet ILSVRC 2012 dataset \cite{russakovsky2015imagenet} were used. The following describes the experiments carried out on each dataset.

\subsection{Effect of the Training Framework on the SoftMax Score Surface for the 2D Points Dataset}
This experiment was carried out to examine the feasibility of training a DL model to learn a SoftMax score surface which stays far from each class distribution similar to the example shown in Fig. \ref{fig:norm_ideal} (right). A simple neural network consisting of 4 fully-connected layers with TanH activation functions was trained to perform binary classification on the 2D points dataset shown in Fig.~\ref{fig:2dpoint} (a). The unlabeled data points generated for this dataset are shown in gray in Fig. \ref{fig:2dpoint} (b). Fig. \ref{fig:2dpoint} (c) shows the SoftMax score surface the model learns without utilizing regularization and unlabeled data. We also show the learned SoftMax score surface by the Jacobian Regularizer which does not use unlabeled data in Fig. \ref{fig:2dpoint} (d). Fig. \ref{fig:2dpoint} (e) shows the SoftMax score surface learned by the model using the naive regularization term and unlabeled data. These results show that the decision boundary learned by the model with normal training gets close to the support of the class distributions at certain locations and the SoftMax score surface is considerably steep around the decision boundary. The SoftMax score surface learned by the model trained by the Jacobian regularization term is slightly smoother than that of the model with normal training. However, there are certain locations outside the support of the data distribution where the SoftMax score surface is steep. The model trained using the naive regularizer and unlabeled data, however, keeps the decision boundary away, far from both class distributions and the SoftMax score surface stays smooth outside the support of the data distribution. These results show that unlabeled data helps the model to learn a SoftMax score surface similar to the ideal SoftMax score surface described despite the fact that a weaker regularization term was used compared to the stronger Jacobian regularizer. More examples of different 2D datasets are provided in Appendix A of the Supplementary Materials.

\subsection{Numerical Results on the Distance of Decision Boundary from the Data Distribution}
\textbf{MNIST Dataset}: The MNIST dataset consists of $60,000$ monochrome images of handwritten digits in the training set. It has 10 classes which correspond to the digits 0 to 9 and an equal number of images per class. The images were resized to $32 \time 32$, rescaled to the range $[0,1]$ and z-score standardized with mean and standard deviation of the training dataset.
\begin{table}[t]
    \centering
    \caption{Column $\mathbf{\hat{\rho}_{adv}}$ shows measured $\mathbf{\hat{\rho}_{adv}}$ values using DeepFool attack for the DL models trained on the MNIST dataset. Test accuracy is also shown for each model. Test accuracy is defined as prediction accuracy of the model on the clean test dataset. Average robust accuracy of the models is also shown against FGSM \cite{goodfellow2014explaining} and PGD \cite{madry2017towards} attacks using epsilon value 0.7.}%
    \resizebox{\columnwidth}{!}{
    \begin{tabular}{cccccc}
    \hline
     & \multirow{2}{*}{Defence method} &  $\mathbf{\hat{\rho}_{adv}}$ & Test accuracy & Robust accuracy & Robust accuracy \\
     & &  $(\times 10^{-2})$ & (\%) & (FGSM, $\epsilon=0.7$) & (PGD, $\epsilon=0.7$)\\ [5pt]
    \hline
     \multirow{8}{*}{Literature}& No defence & 20.67 & 99.08 & -&  -\\
     \cline{2-2}
     & Adversarial training \cite{goodfellow2014explaining} & 22.38 & 99.03 & -& -\\
    \cline{2-2}
     & Input Gradient~\cite{ross2018improving} & 23.43 & \bf{99.25} & -& -\\
    \cline{2-2}
     & Input Gradient~\cite{ross2018improving} \& & \multirow{2}{*}{23.49} & \multirow{2}{*}{98.88} &  \multirow{2}{*}{-}&  \multirow{2}{*}{-}\\
     & Adversarial training \cite{goodfellow2014explaining}& & & &\\
     \cline{2-2}
     & Jacobian \cite{Jakubovitz2018improving} & 34.24 & 98.44 & -& -  \\
    \cline{2-2}
     & Jacobian~\cite{Jakubovitz2018improving} \& & \multirow{2}{*}{36.29} & \multirow{2}{*}{98.00} & \multirow{2}{*}{-}& \multirow{2}{*}{-} \\
     & Adversarial training \cite{goodfellow2014explaining}& & & &\\
       \cline{2-2}
     & TRADES \cite{zhang2019theoretically}  & 25.82 &  99.21 & 86.13 $\mp$ 0.97& 54.90 $\mp$ 3.03\\
        \cline{2-2}
      & TRADES \cite{zhang2019theoretically} \& & \multirow{2}{*}{33.54} & \multirow{2}{*}{99.17}  & \multirow{2}{*}{90.83 $\mp$ 0.33}& \multirow{2}{*}{\bf{67.09 $\mp$ 0.54}}\\
     & Adversarial training \cite{goodfellow2014explaining}& & & & \\
      \hline
    \multirow{5}{*}{Ours}  & No defence &  20.30 & 98.29 & 41.40 $\mp$3.21& 0.02 $\mp$ 0.01\\
      \cline{2-2}
    & Adversarial training \cite{goodfellow2014explaining} & \bf{61.84} & 98.66 & 89.50 $\mp$ 3.30& 43.82 $\mp$ 24.08 \\
      \cline{2-2}
    & Naive regularizer \& Unlabeled data  & 46.29 & 98.69 & 62.62 $\mp$ 2.11& 4.11 $\mp$ 1.07\\
    \cline{2-2}
     & Naive regularizer, Unlabeled data \& & \multirow{2}{*}{51.20}  & \multirow{2}{*}{98.80} & \multirow{2}{*}{\bf{93.62 $\mp$ 1.17}} & \multirow{2}{*}{55.66 $\mp$ 1.87}\\
     & Adversarial training \cite{goodfellow2014explaining}& & & & \\
     \hline
    \end{tabular}}
    \label{tab:mnist}
\end{table}
The LeNet \cite{lecun1998gradient} network architecture and Sigmoid activation function were used for all experiments on the MNIST dataset. Details of the training hyperparameters are given in Appendix B of our Supplementary Materials. Five models were trained to perform classification on the MNIST dataset using our proposed loss function (Eq. \ref{eq:model_loss}). Different values of $\lambda$ were tested and the model's loss curve was checked for each $\lambda$ value. The highest $\lambda$ value that did not decrease the models' validation accuracy by more than 5\% was chosen. We show adversarial robustness of the trained models with same number of models trained using Jacobian Regularization \cite{Jakubovitz2018improving}, Input Gradient Regularization \cite{ross2018improving}, TRADES \cite{zhang2019theoretically} and adversarial training. Table \ref{tab:mnist} shows the results obtained using Jacobian Regularization \cite{Jakubovitz2018improving}, Input Gradient Regularization \cite{ross2018improving}, TRADES \cite{zhang2019theoretically}, adversarial training and our naive regularizer with unlabeled data. The $\hat{\rho}_{adv}$ value defined as $\hat{\rho}_{adv}(f_\theta) = \sum_{x_k \in X}\frac{1}{|X|}\frac{||\Delta x_k||_2}{||x_k||_2}$ measures the average relative distance between data points and the decision boundary and was introduced by Moosavi et al. \cite{moosavi2016deepfool} as a measure of robustness of models against adversarial attacks. Our analysis shows that the  $\hat{\rho}_{adv}$  values were relatively high for the model trained with the naive regularizer and unlabeled data, and when combined with adversarial training compared to other regularization methods. Jacobian Regularization \cite{Jakubovitz2018improving}, Input Gradient Regularization \cite{ross2018improving}, and adversarial training also show larger $\hat{\rho}_{adv}$ value compared to models trained with no defence mechanism. We have also reported the test accuracy of each model in the same table. We define test accuracy as the prediction accuracy of the model on the clean test dataset. The test accuracy achieved by all the models is very close to the test accuracy for models with no defense. Finally, we have also reported the average robust accuracy of the trained DL models against FGSM and PGD attacks for epsilon value 0.7. This epsilon value is the largest epsilon value that DL models trained on MNIST dataset are evaluated for in the literature. We put '-' in the table if we could not find these measurements for those methods which we have taken from literature. Models trained with smaller values than the chosen $\lambda$ value had smaller robust accuracy while models trained with larger values than the chosen $\lambda$ value had smaller test accuracy. 

\textbf{CIFAR10 Dataset}: The CIFAR10 dataset contains color images of 10 natural image classes with equal numbers of images per class.
\begin{table}[t]
    \centering
    \caption{Column $\mathbf{\hat{\rho}_{adv}}$ shows measured $\mathbf{\hat{\rho}_{adv}}$ values using DeepFool attack for the models trained on CIFAR10 dataset. Test accuracy is also shown for each model. Test accuracy is defined as prediction accuracy of the model on the clean test dataset. Average robust accuracy of the models is also shown against FGSM \cite{goodfellow2014explaining} and PGD \cite{madry2017towards} attacks using epsilon value 15.}
     \resizebox{\columnwidth}{!}{
    \begin{tabular}{cccccc}
    \hline
     & \multirow{2}{*}{Defence method} &  $\mathbf{\hat{\rho}_{adv}}$ & Test accuracy & Robust accuracy & Robust accuracy \\
     & &  $(\times 10^{-2})$ & (\%) & (FGSM, $\epsilon=15$) & (PGD, $\epsilon=15$)\\[5pt]
     \hline
      \multirow{8}{*}{Literature}& No defence &  1.21 & 88.79 & -& -\\
     \cline{2-2}
     & Adversarial training \cite{goodfellow2014explaining}& 1.23 & 88.88 &  -&  -\\
    \cline{2-2}
     & Input Gradient~\cite{ross2018improving} &  1.43 & 88.56 &  - &  -\\
      \cline{2-2}
     & Input Gradient~\cite{ross2018improving} \&  & \multirow{2}{*}{2.17} & \multirow{2}{*}{88.49} &  \multirow{2}{*}{-} &  \multirow{2}{*}{-}\\
     & Adversarial training \cite{goodfellow2014explaining} & & & & \\
     \cline{2-2}
     & Jacobian \cite{Jakubovitz2018improving} & 3.42 & 89.16 &  - & -\\
     \cline{2-2}
     & Jacobian \cite{Jakubovitz2018improving}\& & \multirow{2}{*}{\bf{6.03}} & 
     \multirow{2}{*}{88.49} &  \multirow{2}{*}{-}& \multirow{2}{*}{-}\\
     & Adversarial training \cite{goodfellow2014explaining} & & & &\\
        \cline{2-2}
     & TRADES \cite{zhang2019theoretically}  & 1.44 &  86.74 &  16.11 $\mp$ 0.75 &  7.74 $\mp$ 0.28\\
        \cline{2-2}
      & TRADES \cite{zhang2019theoretically} \& & \multirow{2}{*}{1.34} & \multirow{2}{*}{87.15} & \multirow{2}{*}{16.78 $\mp$ 1.30} &   \multirow{2}{*}{7.99 $\mp$ 0.57} \\
     & Adversarial training \cite{goodfellow2014explaining}& & & &\\
     \hline
      \multirow{6}{*}{Ours} & No defence & 1.23 & \bf{91.17} & 10.61 $\mp$ 0.39 & 1.49 $\mp$ 0.08\\
      \cline{2-2}
       & Adversarial training \cite{goodfellow2014explaining}& 1.27 & 89.51 & \bf{80.06 $\mp$ 1.74} & 2.30 $\mp$ 0.89\\
      \cline{2-2}
      &   Naive regularizer \& Unlabeled data &  3.29 & 83.96 & 47.03 $\mp$ 0.85 & 43.60 $\mp$ 0.67\\
     \cline{2-2}
     &  Naive regularizer, Unlabeled data \& &  \multirow{2}{*}{4.45}  &  \multirow{2}{*}{83.50} & \multirow{2}{*}{64.49 $\mp$ 3.22} & \multirow{2}{*}{\bf{58.46 $\mp$ 2.78}}\\
     & Adversarial training \cite{goodfellow2014explaining} & & & &\\
     \hline
    \end{tabular}}
    \label{tab:cifar}
\end{table}
CIFAR10 images were also rescaled to the range $[0,1]$ and z-score standardized using the mean and standard deviation of the training dataset and augmentation methods were applied during training. For CIFAR10 dataset, ResNet 9 \cite{he2016deep} network architecture was used with a CELU activation function for all layers. The lambda value was chosen similarly as for the MNIST dataset. Details of the training hyperparameters are given in Appendix B of our Supplementary Materials. Five models were trained using the proposed loss function (Eq. \ref{eq:model_loss}), the labeled dataset and the generated unlabeled dataset. 
Training and comparison of the results were conducted similar to the MNIST experiments. Table \ref{tab:cifar} shows the results obtained using Jacobian Regularization \cite{Jakubovitz2018improving}, Input Gradient Regularization \cite{ross2018improving}, TRADES \cite{zhang2019theoretically} and the naive regularizer with unlabeled data. The $\hat{\rho}_{adv}$ values achieved for the naive regularizer and unlabeled data and when combined with adversarial training are larger compared to models trained with no defense, adversarial training, Input Gradient Regularization \cite{ross2018improving} and TRADES \cite{zhang2019theoretically}. Jacobian regularization and Jacobian Regularization with adversarial training show larger  $\hat{\rho}_{adv}$  values. The test accuracy achieved was slightly lower using our training framework. We believe the test accuracy is compromised over achieving a smooth SoftMax surface outside the support of the data distribution in the high dimensional input data space. Also shown is the average robust accuracy of the trained DL models against FGSM and PGD attacks for epsilon value 15. We put '-' in the table if we could not find these measurements for those methods which we have taken from literature.

\textbf{Imagenette Dataset}: Imagenette is a subset of the well-known ImageNet ILSVRC 2012 dataset \cite{russakovsky2015imagenet} with 10 easily classified classes. Images were rescaled to the range $[0,1]$ and z-score standardized using the mean and standard deviation of images in the training dataset. The XResNet 18 \cite{he2019bag} network architecture was used to train a model to classify Imagenette dataset with MISH activation function. Details of the training hyperparameters are given in Appendix B of our Supplementary Materials.
\begin{table}[t]
    \centering
    \caption{Column $\mathbf{\hat{\rho}_{adv}}$ shows measured $\mathbf{\hat{\rho}_{adv}}$ values using DeepFool attack for the models trained on Imagenette dataset. Test accuracy is also shown for each model. Test accuracy is defined as prediction accuracy of the model on the clean test dataset. Average robust accuracy of the models is also shown against FGSM \cite{goodfellow2014explaining} and PGD \cite{madry2017towards} attacks using epsilon value 19.}
     \resizebox{\columnwidth}{!}{
    \begin{tabular}{cccccc}
    \hline
     &\multirow{2}{*}{Defence method} &  $\mathbf{\hat{\rho}_{adv}}$ & Test accuracy & Robust accuracy & Robust accuracy \\
     & &  $(\times 10^{-2})$ & (\%) & (FGSM, $\epsilon=19$, \%) & (PGD, $\epsilon=19$, \%)\\[5pt]
    \hline
     \multirow{6}{*}{Literature} & 
     Jacobian \cite{Jakubovitz2018improving} & 2.49 &  85.60 & 23.67 $\mp$ 0.68& 8.54 $\mp$ 0.48  \\
    \cline{2-2}
     & Jacobian~\cite{Jakubovitz2018improving} \& & \multirow{2}{*}{4.39} & \multirow{2}{*}{\bf{87.18}} & \multirow{2}{*}{42.71 $\mp$ 0.47}& \multirow{2}{*}{28.21 $\mp$ 0.36} \\
     & Adversarial training \cite{goodfellow2014explaining}& & & &\\
    \cline{2-2}
     & TRADES \cite{zhang2019theoretically}  & 1.20 &  85.97 &  7.02 $\mp$ 0.23 &   0.08 $\mp$ 0.01\\
    \cline{2-2}
     & TRADES \cite{zhang2019theoretically} \& & \multirow{2}{*}{4.15} & \multirow{2}{*}{82.88} & \multirow{2}{*}{56.08 $\mp$ 0.40} & \multirow{2}{*}{53.18 $\mp$ 0.39} \\
     & Adversarial training \cite{goodfellow2014explaining}& & & &\\
     \hline
    \multirow{5}{*}{Ours}& No defence & 0.941 & \bf{87.18}  & 6.48 $\mp$  0.37 & 0.01 $\mp$ 0.00\\
     \cline{2-2}
    & Adversarial training \cite{goodfellow2014explaining}& 4.82 & 83.06 & \bf{69.86 $\mp$  0.45}&   67.22 $\mp$ 0.52\\
    \cline{2-2}
    &  Naive regularizer \& Unlabeled data & 2.21 & 84.92 & 12.69 $\mp$ 0.79 & 0.55 $\mp$ 0.09\\
    \cline{2-2}
    &  Naive regularizer, Unlabeled data \&  & \multirow{2}{*}{\bf{4.96}} & \multirow{2}{*}{82.44} & \multirow{2}{*}{69.59 $\mp$ 0.39} & \multirow{2}{*}{\bf{67.68 $\mp$ 0.37}}\\
    & Adversarial training \cite{goodfellow2014explaining}&  & & & \\
     \hline
    \end{tabular}}
    \label{tab:imagenette}
\end{table}
The training and attacks performed were similar to those in the experiments for MNIST and CIFAR10 datasets. Table \ref{tab:imagenette} shows the results obtained without regularization, with adversarial training, and the naive regularizer with unlabeled data. Also shown is the results obtained using Jacobian regularizer \cite{Jakubovitz2018improving} and TRADES \cite{zhang2019theoretically} with adversarial training. Similar to the previous experiments, our models show higher $\hat{\rho}_{adv}$ values compared to normal training and TRADEs \cite{zhang2019theoretically}. When combined with adversarial training, the model shows the highest $\hat{\rho}_{adv}$ value. \par
The experiments on MNIST, CIFAR10 and Imagenette datasets show that a naive regularizer together with unlabeled data indeed results in models with larger $\hat{\rho}_{adv}$, which implies that the distance between data points and decision boundary increases for the models trained with unlabeled data that supports a larger area in the high dimensional input data space. We showed that even a naive regularizer can improve adversarial robustness of DL models using our generated unlabeled data. We defined and used a naive regularizer so that it would allow us to run rigorous experiments to study our hypotheses in sections \ref{sec:unlabeled} to \ref{sec:uuc}. Due to the slow speed of the Jacobian regularizer, we were not able to use this regularizer to run the experiments in sections \ref{sec:unlabeled} to \ref{sec:uuc} in a reasonable amount of time. However, we experimentally show Jacobian regularizer’s superiority over our naive regularizer in section \ref{sec:comparison_with_jacobian_regularization}. We believe other regularization techniques such as Input Gradient Regularization \cite{ross2018improving} and Jacobian Regularization \cite{Jakubovitz2018improving} and adversarial training are indirectly trying to achieve the ideal decision boundary explained before. The drawback of these algorithms, however, is that they do not take advantage of unlabeled data during training. In section \ref{sec:comparison_with_jacobian_regularization}, we conduct experiments to analyse how models trained with Jacobian regularization and unlabeled data perform in comparison to Jacobian regularization alone and the naive regularizer with unlabeled data. In other words, does unlabeled data help the models trained with Jacobian regularization perform even better. In the next section, we show the results of these experiments.

\subsection{Comparison between Our Regularization and the Jacobian Regularization}
\label{sec:comparison_with_jacobian_regularization}
As mentioned before, applying Jacobian regularization during the training process is very time consuming. However, this regularizer is stronger than our naive regularization term. For the purpose of confirming this point, we ran multiple experiments and compared the robust accuracy of models trained with the Jacobian regularization term and the naive regularization term. Although Jakubovitz and Giryes \cite{Jakubovitz2018improving} applied Jacobian regularization as a post-processing step after training the model has finished, we applied the Jacobian regularizer to the model during the training process similar to our training framework to ensure a fair comparison. We compared the robustness of models trained with the naive regulaizer and unlabeled data, the Jacobian regularizer without unlabeled data and the Jacobian regularization with unlabeled data using the FGSM \cite{goodfellow2014explaining} and PGD \cite{madry2017towards} attacks. \par

\begin{figure}[t]
    \centering
     \includegraphics[clip=true, trim=4cm 2cm 4cm 0cm,width=\textwidth]{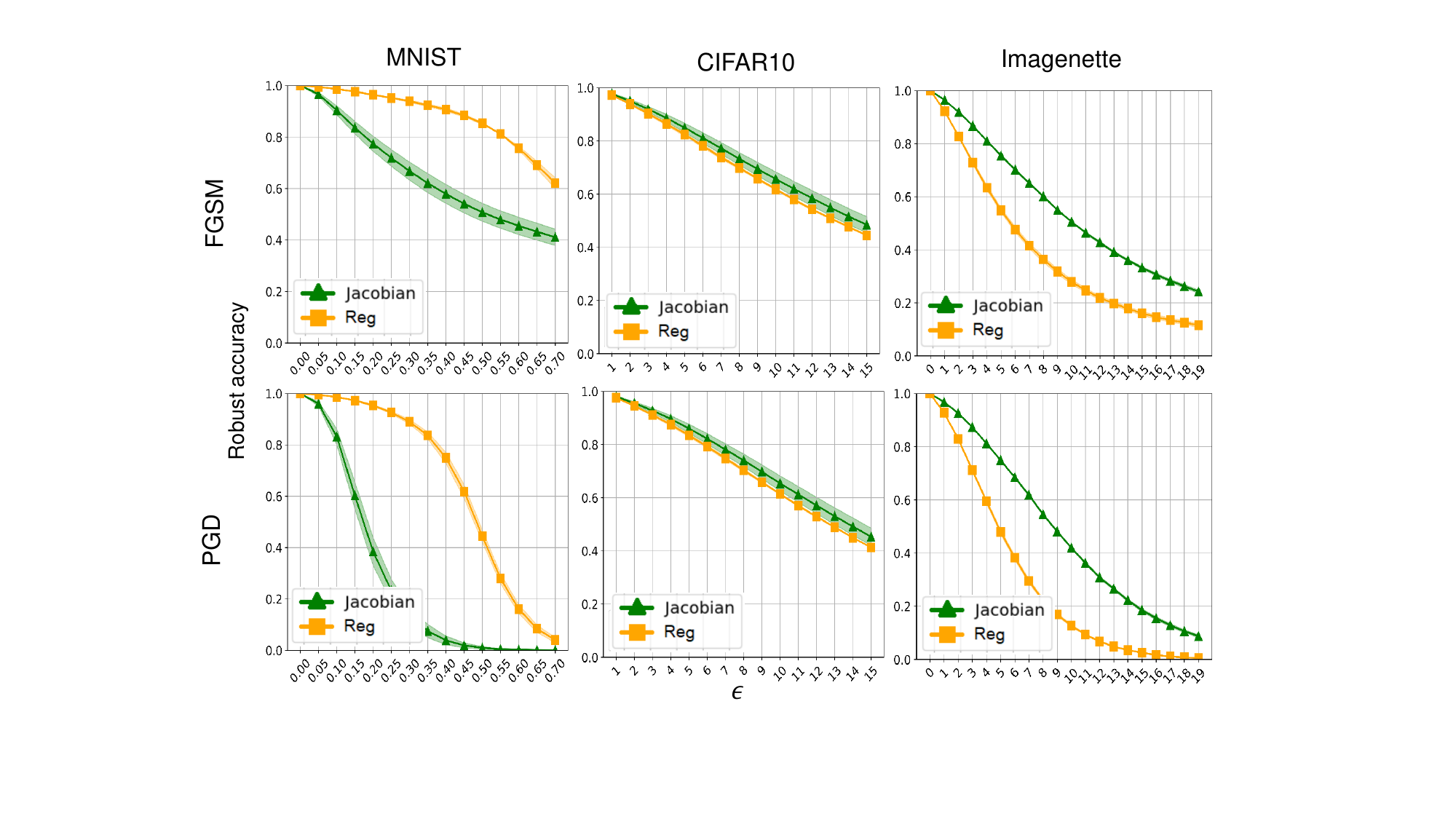}
    \caption{Figure shows the robust accuracy of models trained using the naive regularizer (Reg) with unlabeled data and the Jacobian regularization (Jacobian) without unlabeled data \cite{Jakubovitz2018improving} for MNIST, CIFAR10 and Imagenette datasets. The robust accuracy is measured using FGSM \cite{goodfellow2014explaining}, and PGD \cite{madry2017towards} attacks with different epsilon perturbation values ($\epsilon$). The reader is referred to the online version of this paper for the color representation of this figure.}
    \label{fig:jacobianreg}
\end{figure}

For each dataset, three models were trained using the naive regularizer and unlabeled data and three models were trained using the Jacobian regularizer without unlabeled data. Their robust accuracy evaluated against the FGSM \cite{goodfellow2014explaining} and PGD \cite{madry2017towards} attacks with different epsilon perturbation values ($\epsilon$) is outlined in Figure \ref{fig:jacobianreg}. The lambda values used for the Jacobian Regularizer were set to 0.01 for the MNIST dataset, 0.01 for the CIFAR10 dataset and 0.005 for the Imagenette dataset and were found as specified in the previous section. \par

As shown in Figure \ref{fig:jacobianreg}, the proposed training framework performs better for the MNIST dataset, whereas the Jacobian regularizer performs better for the CIFAR10 and Imagenette dataset. This improved performance has two potential explanations: 
\begin{enumerate}
\item The data sparsity hypothesis is proven wrong for the CIFAR10 and Imagenette datasets. 
\item The Jacobian regularizer is a stronger regularizer than our naive regularizer, regardless of the extent to which the data distribution supports the high dimensional input data space. 
\end{enumerate}
To find out which explanation is true, we compared the robustness of models trained using just the Jacobian regularizer versus models trained with both Jacobian regularization and unlabeled data. Three models were trained per dataset. If the robustness does not significantly improve after the Jacobian regularizer was deployed with unlabeled data, we could conclude that unlabeled data does not provide significant benefits for training models on CIFAR10 and Imagenette datasets. In other words, the first explanation holds true. Conversely, if the robustness does significantly improve, the second explanation holds true. \par
\begin{figure}[t]
    \centering
     \includegraphics[clip=true, trim=4cm 2.2cm 4.2cm 0.2cm,width=1\textwidth]{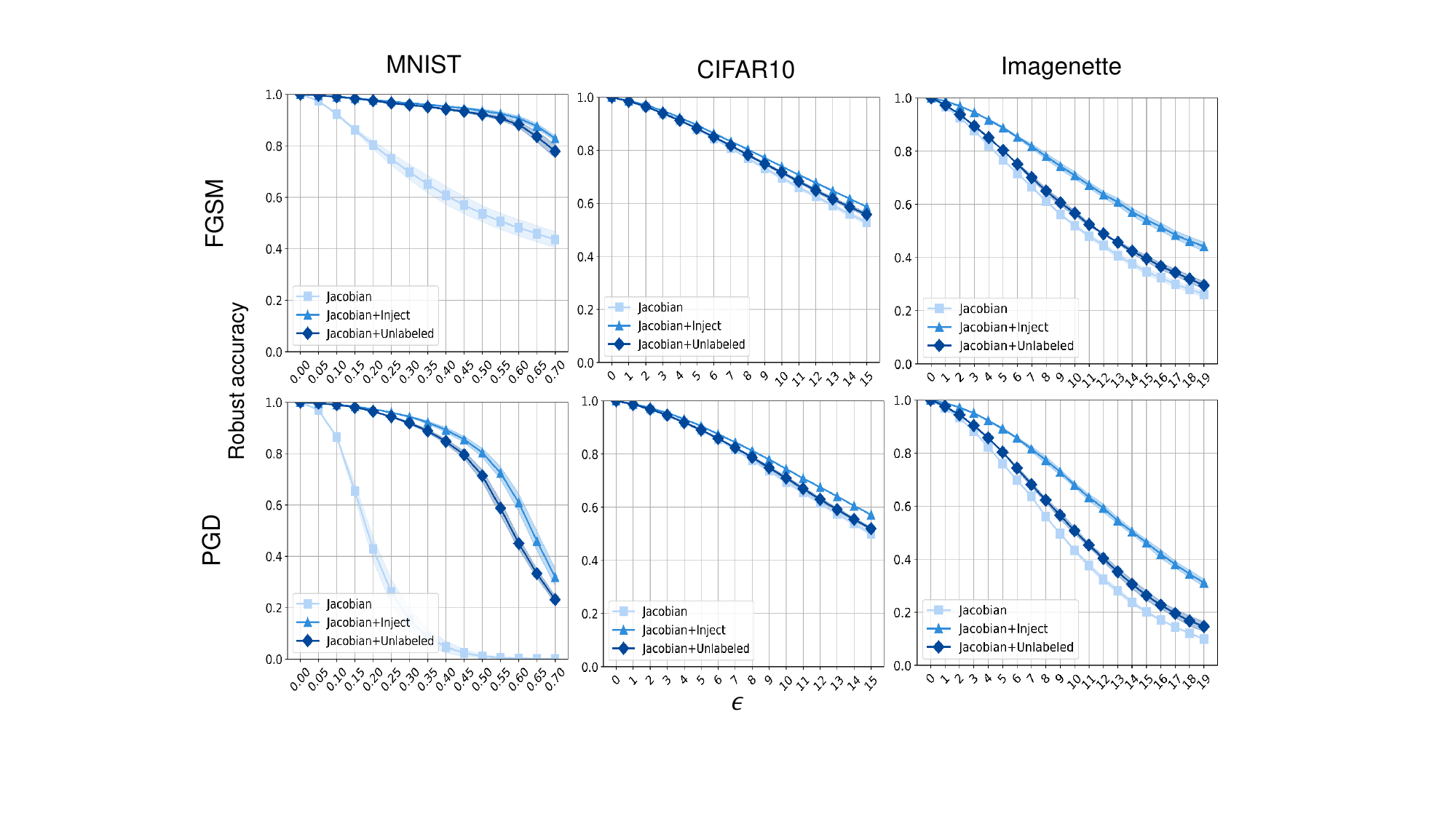}
    \caption{Figure shows the robust accuracy of models trained using the Jacobian regularization term (Jacobian) versus models trained using both Jacobian regularization and unlabeled data (Jacobian+Unlabeled) and the Jacobian regularization with Gaussian noise injected to the intermediate model outputs (Jacobian+Inject). The robust accuracy is measured using FGSM \cite{goodfellow2014explaining} and PGD \cite{madry2017towards} attacks with different epsilon perturbation values ($\epsilon$). It can be seen that training models using both Jacobian regularization and unlabeled data and Jacobian regularization with injected Gaussian noise to the intermediate model outputs are more resilient to adversarial perturbations. The reader is referred to the online version of this paper for the color representation of this figure.}
    \label{fig:jacobian_vs_unlabeled}
\end{figure}

\begin{table*}[t]
\centering
\caption{Columns $\mathbf{\hat{\rho}_{adv}}(\times10^{-2})$ and Test Accuracy (\%) show measured $\mathbf{\hat{\rho}_{adv}}$ values from the DeepFool attack \cite{moosavi2016deepfool} and test accuracy for the models trained on MNIST, CIFAR10, and Imagenette datasets. This table compares the robustness of models trained using the Jacobian regularization (Jacobian), Jacobian regularization and unlabeled data (Jacobian+Unlabeled), as well as Jacobian regularization with Gaussian noise injected to the intermediate model outputs (Jacobian+Inject).}
\resizebox{\columnwidth}{!}{%
\begin{tabular}{ccccccc}
 & \multicolumn{2}{c}{MNIST} & \multicolumn{2}{c}{CIFAR10} & \multicolumn{2}{c}{Imagenette} \\ \hline
 \multirow{2}{*}{Defence method} & $\hat{\rho}_{adv}$ & Test accuracy & $\hat{\rho}_{adv}$ & Test accuracy & $\hat{\rho}_{adv}$ & Test accuracy \\
 & $(\times 10^{-2})$ &(\%) & $(\times 10^{-2})$ & (\%) & $(\times 10^{-2})$ & (\%) \\ \hline
Jacobian & 561 & 98.75 & 3.56 & 88.01 & 2.57 & 85.52 \\ \hline
Jacobian+Unlabeled & 1083 & 98.84 & 3.98 & 87.78 & 2.73 & 85.29 \\ \hline
Jacobian+Inject & 1582 & 98.70 & 4.42 & 87.24 & 4.21 & 86.93 \\ \hline
\end{tabular}}
\label{tab:jacobianrhos}
\end{table*}
The $\hat{\rho}_{adv}$ values and test accuracy of the trained models are shown in the rows "Jacobian", and "Jacobian+Unlabeled" in Table \ref{tab:jacobianrhos}. The robust accuracy is also shown with labels "Jacobian" and "Jacobian+Unlabeled" in Figure \ref{fig:jacobian_vs_unlabeled}. As seen in Table \ref{tab:jacobianrhos} and Figure \ref{fig:jacobian_vs_unlabeled}, when the Jacobian regularizer is used in conjunction with unlabeled data, robustness indeed improves. Therefore, the second explanation holds true. Jacobian Regularization is indeed more powerful than our naive regularizer. The proposed naive regularizer uses neighbor points to minimize the slope of the SoftMax score surface. Depending on the relative position of the neighbor data point with respect to the labeled or unlabeled data point, our naive regularizer smooths the SoftMax score surface in not all but a limited number of dimensions in the input data space in exchange for a shorter training runtime whereas the Jacobian regularizer smooths the SoftMax score surface in all dimensions in the input data space at the location of the input data point. A comparison of the runtime for the naive regularizer with unlabeled data, the Jacobian regularizer and the Jacobian regularizer with unlabeled data is shown in Table \ref{tab:runtime} where the average training runtime for 1 epochs is reported in minutes for each dataset. One can see that the runtime for the naive regularizer with unlabeled data is less than the training time for the Jacobian regularizer with unlabeled data by 1.5-4 times. 

\begin{table}[t]
    \centering
     \caption{Table shows the average runtime (in minutes) for training a LeNet network \cite{lecun1998gradient} with MNIST dataset for 1 epoch, the runtime for training a ResNet 9 network \cite{he2016deep} on CIFAR10 dataset and a XResNet 18 network \cite{he2019bag} on Imagenette dataset using our naive regularization term with unlabeled data, the Jacobian regularization term and the Jacobian regularization term with unlabeled data. All measurements were done on a Google Colab T4 Tesla GPU for a single run.}
    \resizebox{\columnwidth}{!}{%
    \begin{tabular}{cccc}
    \hline
       Training runtime & Proposed + Unlabeled (min) & Jacobian (min)& Jacobian + Unlabeled (min)\\
       \hline
        MNIST &5.82 & 4.05 & 8.72\\ 
        \hline
        CIFAR10 & 11.83 & 12.47 & 25.75\\
        \hline
        Imagenette &  8.48 & 16.80 & 34.55\\
        \hline
    \end{tabular}}
    \label{tab:runtime}
\end{table}

The reader may find certain differences in the robust accuracy plots in our paper and the plots in the literature.There are two reasons for the differences seen in the plots in our paper and the plots in the literature. The first reason is that unlike the papers in the literature where robust accuracy is reported for all samples in the test dataset, we show adversarial robustness of a model for samples that the model originally classified correctly. In other words, we only perturb images that the model classifies correctly and see if the model can still assign the perturbed images to the correct class label. We believe showing how much a model’s performance deteriorates on correctly classified images is a more accurate measurement of robust accuracy of models. \par

Another reason for the difference seen in the plots can be due to differences in the data preprocessing pipeline of our experiments and the experiments conducted in the literature. In our preprocessing pipelines, we followed standard preprocessing steps in the literature for the three datasets: MNIST, CIFAR10, and Imagenette. We z-score standardized image intensities using the mean and standard deviation of all the images in the training dataset. We also applied augmentation methods for CIFAR10 and Imagenette datasets. The preprocessing pipelines are included in our code that is available online. However, we were not able to find out what preprocessing steps were performed for the algorithms in the literature such as Input Gradient \cite{ross2018improving} and Jacobian \cite{Jakubovitz2018improving} regularization algorithms. We used the standard Foolbox library \cite{rauber2017foolboxnative} to measure robust accuracy for our models. In the next section, the effect of different types of unlabeled data on the adversarial robustness of DL models is analysed.

\begin{figure}[t] 
    \centering
   \includegraphics[clip=true, trim=0cm 10.2cm 0.1cm 0cm,width=\textwidth]{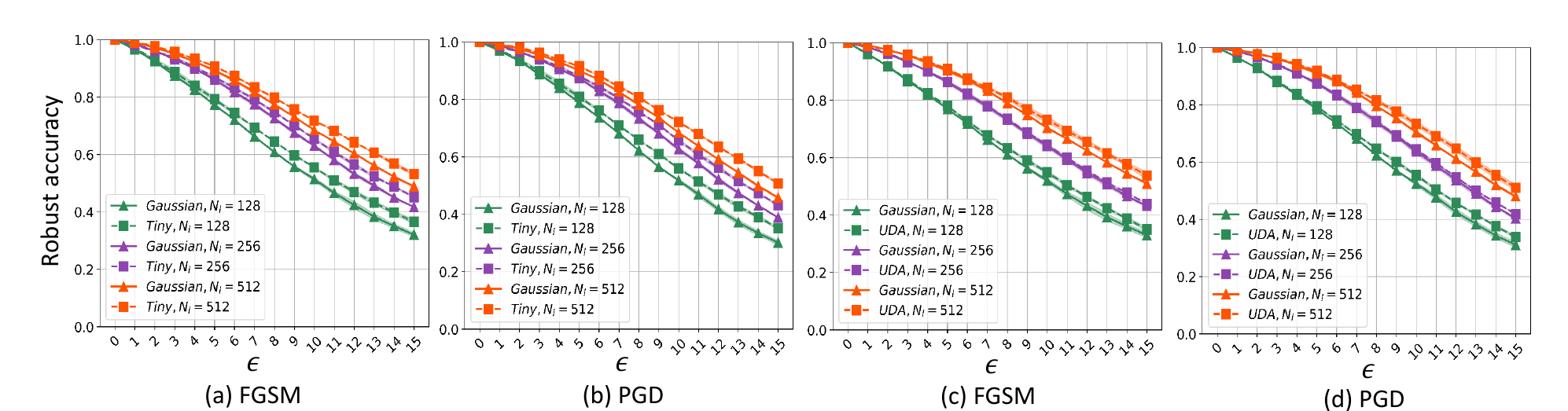}\\
    \caption{Figure shows robustness of two models trained with regularization using a Gaussian generated dataset as the unlabeled data and using the 80m@$200k$ dataset as the unlabeled data against FGSM (a) and PGD (b) attacks for models trained with 128, 256, and 512 labeled data points per class ($N_l$) from the CIFAR10 dataset. (c) and (d) show robustness of two models trained with regularization using a Gaussian generated dataset as the unlabeled and using the UDA generated dataset \cite{xie2019unsupervised} as the unlabeled data against FGSM and PGD attacks respectively. The reader is referred to the online version of this paper for the color representation of this figure.}
    \label{fig:gaussian_vs_tiny_vs_uda}
\end{figure}
\subsection{Gaussian Noise is as Effective as Costly and More Complex Unlabeled Datasets}
\label{sec:unlabeled}
An important question is whether models trained with unlabeled datasets widely utilized in the literature for SSL show a higher adversarial performance compared to models trained with unlabeled data generated with Gaussian noise. To answer this question we trained two models for CIFAR10 dataset using our naive regularizer with unlabeled data extracted from the 80 Million Tiny Images dataset following the same automatic filtering technique used by Xie et al. \cite{xie2019unsupervised}. We also prepared another unlabeled dataset by applying the synthesis technique proposed by Uesato et al.\cite{uesato2019are}. Similar to many papers on semi-supervised learning and regularization \cite{miyato2018virtual, springenberg2015unsupervised, zhai2019adversarially}, we ran experiments for different numbers of labeled data points per class ($N_l = 128, 256, 512$) and the same number of unlabeled data points per labeled data point ($N_u = 1$) using these two unlabeled datasets, and compared the robust accuracy of the trained models with those of models trained with unlabeled data generated from Gaussian noise. Fig. \ref{fig:gaussian_vs_tiny_vs_uda} (a) and (b) compare robust accuracy of models trained with unlabeled data extracted from the 80 Million Tiny Images dataset and our models using FGSM and PGD attacks. Similarly, Fig \ref{fig:gaussian_vs_tiny_vs_uda} (c) and (d) compare robust accuracy of models trained with unlabeled datasets generated using the synthesis technique proposed in the work by Uesato et al.\cite{uesato2019are} with our models for FGSM and PGD attacks. Our models show slightly less robust accuracy for the same number of unlabeled data points, suggesting that generating unlabeled datasets using Gaussian noise as explained in this paper is almost as effective as more complex and sometimes expensive unlabeled datasets.
\begin{figure}[t]
    \centering
    \includegraphics[clip=true, trim=0.5cm 0.5cm 6.5cm 1cm,width=\textwidth]{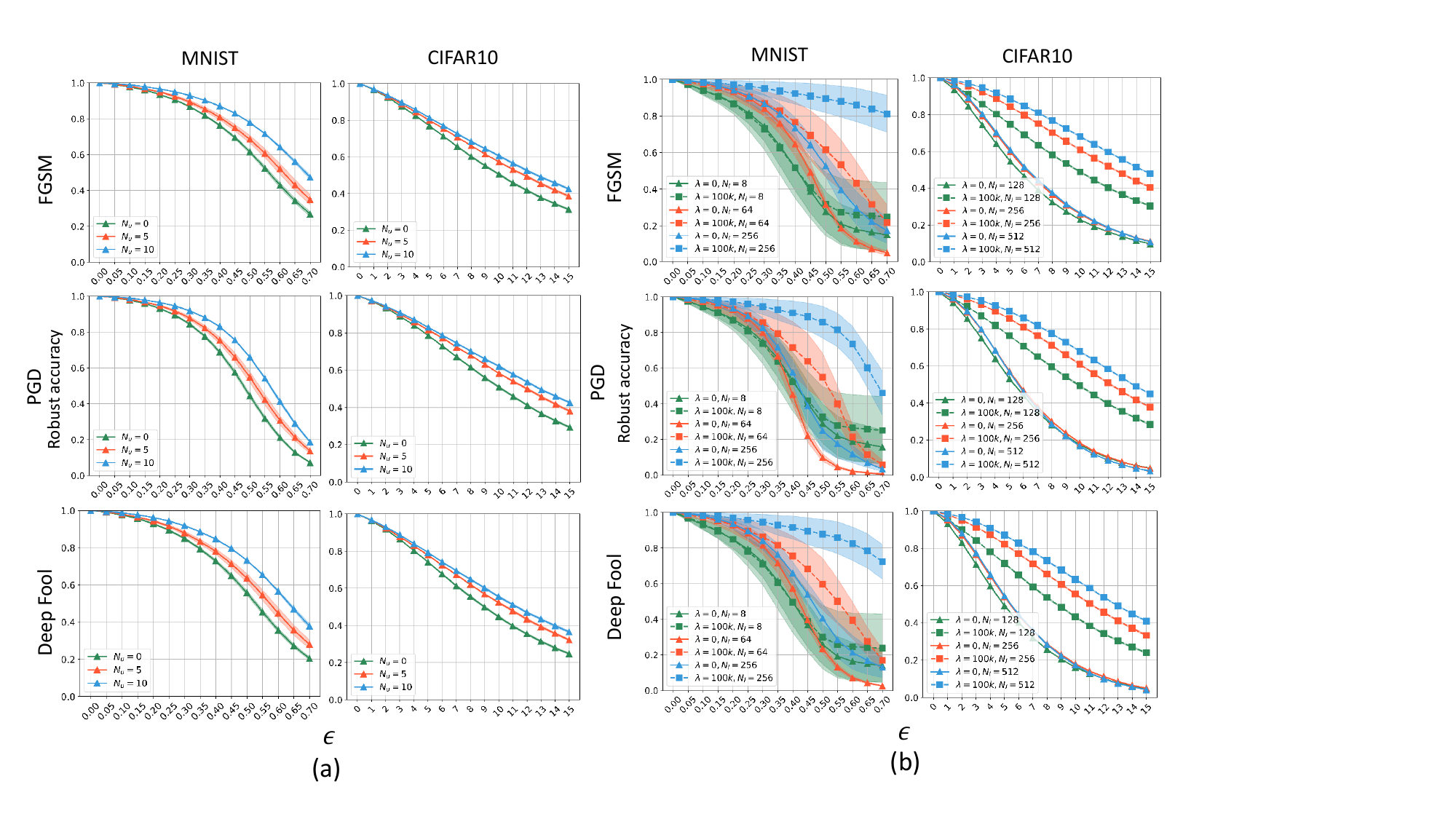} 
    \caption{(a) Shows the robust accuracy for models trained on the MNIST dataset using regularization with $\lambda = 100k$ and $N_l = 64$ (left), and robust accuracy for models trained on the CIFAR10 dataset using regularization with $\lambda=10k$ and $N_l = 128$ for different numbers of unlabeled data points per labeled data points, $N_{u}$ using FGSM, PGD, and DeepFool $(l^{2})$ attacks (right). (b) Shows the robust accuracy for models trained on the MNIST dataset with and without regularization ($\lambda = 100k, 0$)  with different numbers of labeled data points, $N_l = 8, 64, 256$ (left). (Right) Shows the robust accuracy for models trained on the CIFAR10 dataset with and without regularization ($\lambda = 10k, 0$) with different numbers of labeled data points, $N_l$, using FGSM, PGD, and DeepFool $(l^{2})$ attacks. The reader is referred to the online version of this paper for the color representation of this figure.}
    \label{fig:lambda_unlabeled_exp}
\end{figure}

\subsection{Effect of Regularization and Unlabeled Data on the Trained DL Models}
This section analyses the effect of our naive regularizer and the amount of unlabeled data on adversarial robustness of DL models. Fig. \ref{fig:lambda_unlabeled_exp}(a) (Left) shows robust accuracy for models trained with different number of unlabeled data per labeled data points ($N_u$ = 0, 5, and 10) using our training framework ($\lambda=100k$), $N_l=64$, for the MNIST dataset while other hyperparameters were fixed. Five models were trained for each set of parameters. Three different attack methods (FSGM, PGD, and DeepFool ($l^{2}$) were conducted on the models, and the mean and standard error of the robust accuracy were measured. The plots show that models trained with larger $N_u$ show higher robust accuracy. Models trained with and without our naive regularizer ($\lambda=100k, 0$) were also compared for different numbers of labeled data points per class ($N_l = 8, 64, 256$) with other hyperparameters fixed. Five different models were trained similarly per set of parameters. All trained models had similar validation accuracy. More details on the experiment design and hyperparameters are provided in Appendix C of the Supplementary Materials. Fig. \ref{fig:lambda_unlabeled_exp} (b) (Left) shows the mean and standard error of the robust accuracy values measured for the trained models suggesting that models trained with our regularizer were more robust compared to models trained without our regularizer. We ran similar experiments on the CIFAR10 dataset. Fig. \ref{fig:lambda_unlabeled_exp} (a) (Right) shows the robust accuracy of models trained with $N_u =$ 1, 5, and 10, $\lambda=10k$, $N_l=128$ with other hyperparameters fixed for the CIFAR10 dataset. All trained models had similar validation accuracy. As shown, the models trained with larger $N_u$ show higher robust accuracy. Fig. \ref{fig:lambda_unlabeled_exp} (b) (Right) shows the mean and standard error of the robust accuracy values measured for models trained with and without our regularizer ($\lambda=10k, 0$) and different number of labeled data points per class ($N_l = 128, 256, 512$) with other hyperparameters fixed. For all three parameter settings, models trained with our naive regularizer were more robust than models trained without our naive regularizer. The experiments show that regularization and unlabeled data both have positive influence on adversarial robustness of trained models. We only show the adversarial robustness of the models using DeepFool ($l^{2}$) attack because the models show a very small decrease in adversarial robustness when attacked with DeepFool ($l^\infty$).

\subsection{The Effect of the Network Layer where Gaussian Noise is Applied}
\begin{figure}[t]
    \centering
    \includegraphics[clip=true, trim=4cm 3.5cm 5.2cm 0.2cm,width=\textwidth]{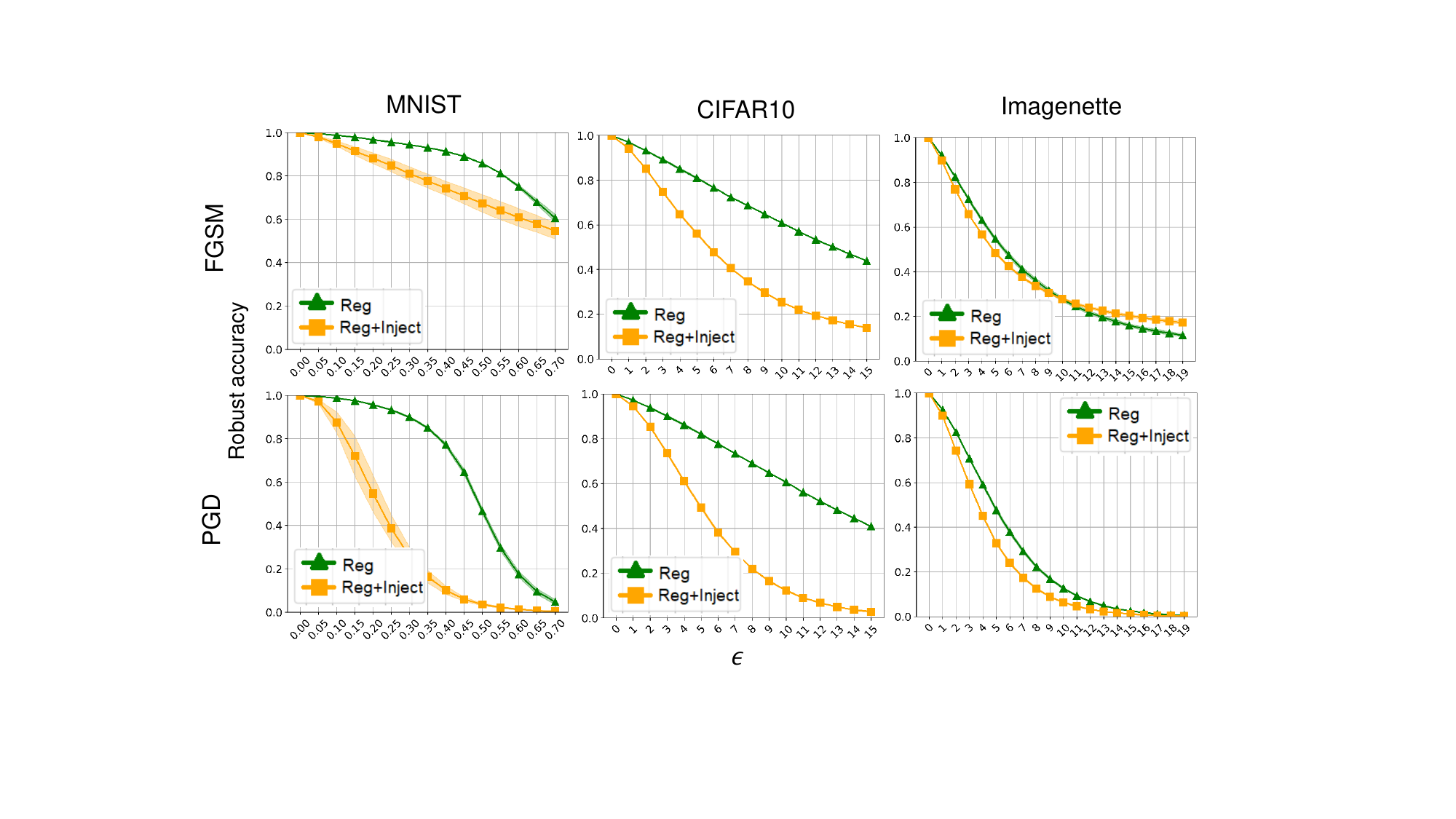}
    \caption{Figure shows the robust accuracy of models trained using the naive regularizer with unlabeled data (Reg) and the naive regularizer with noise injection (Reg+Inject). The robust accuracy is measured using FGSM \cite{goodfellow2014explaining} and PGD \cite{madry2017towards} attacks with different epsilon perturbation values ($\epsilon$). The reader is referred to the online version of this paper for the color representation of this figure.}
    \label{fig:regvsinject}
\end{figure}

In the previous experiments, the unlabeled data was generated once and passed through the model for regularization.  In this section, we conduct experiments to ascertain if applying Gaussian noise to intermediate layers in the network would train models with almost the same adversarial robustness or training models with the generated unlabeled data results in more adversarially robust models. Noise injection in the intermediate layers was conducted in the same way the unlabeled dataset was generated. The mean pair-wise Euclidean distance ($\mu_{pair}$) between the outputs of the previous layer in the network was measured for the labeled dataset and multiples of $\mu_{pair}$ was used as the standard deviation of the Gaussian distribution from which noise was sampled. We trained three models with intermediate noise injection explained in detail in the Appendix D of the Supplementary Materials for each dataset. Table \ref{tab:injectrhos} shows the measured $\hat{\rho}_{adv}$ values, and test accuracy for these models and Fig. \ref{fig:regvsinject} shows the robust accuracy for these models compared with models trained with our naive regularizer and unlabeled data.
\begin{table}[t]
\centering%
\caption{Columns $\hat{\rho}_{adv}(\times10^{-2})$ and Test Accuracy (\%) show measured $\hat{\rho}_{adv}$ values using DeepFool attack \cite{moosavi2016deepfool} and test accuracy for the models trained on the MNIST, CIFAR10, and Imagenette datasets. This table compares the robustness of models trained using the naive regularizer with unlabeled data (Reg + Unlabeled) with models trained using the naive regularizer and intermediate noise injection (Reg + Inject).}
\resizebox{0.9\columnwidth}{!}{%
\begin{tabular}{ccccccc}
 & \multicolumn{2}{c}{MNIST} & \multicolumn{2}{c}{CIFAR10} & \multicolumn{2}{c}{Imagenette} \\ \hline
 \multirow{2}{*}{Defence method} & \multirow{2}{*}{$\hat{\rho}_{adv}$} & Test & \multirow{2}{*}{$\hat{\rho}_{adv}$} & Test  & \multirow{2}{*}{$\hat{\rho}_{adv}$} & Test  \\
 & & accuracy &  & accuracy & & accuracy \\
 & $(\times 10^{-2})$ &(\%) & $(\times 10^{-2})$ & (\%) & $(\times 10^{-2})$ & (\%) \\\hline
Reg + Unlabeled & 46.29 & 98.69 & 3.29 & 83.96 & 2.21 & 84.92 \\ \hline
Reg + Inject & 23.51 & 98.41 & 1.39 & 91.02 & 2.29 & 86.43 \\ \hline
\end{tabular}}
\label{tab:injectrhos}
\end{table}
The results in Table \ref{tab:injectrhos} show that the $\hat{\rho}_{adv}$ values are lower when noise is injected to the intermediate layers of the models during training compared to when models are trained with unlabeled data for the MNIST and CIFAR10 datasets. The $\hat{\rho}_{adv}$ value is higher when noise is injected to the intermediate layers of the models during training for the Imagenette dataset. Figure \ref{fig:regvsinject} also shows that the robust accuracy of models trained with noise injection decreases faster than models trained with unlabeled data when epsilon perturbation value, $\epsilon$, increases except for models trained on the Imagenette dataset and evaluated against FGSM attack. 

We also compared models trained with Jacobian regularization and unlabeled data and Jacobian regularization and intermediate noise injection to investigate if the adversarial robustness would similarly decrease faster for models trained with intermediate noise injection as observed for our naive regularizer. Three models were trained for each dataset in the same way. The $\hat{\rho}_{adv}$ values and test accuracy is shown in Table \ref{tab:jacobianrhos} with labels "Jacobian + Unlabeled" and "Jacobian + Inject", respectively. The corresponding robust accuracy is also shown in Figure \ref{fig:jacobian_vs_unlabeled}. The results show models trained with intermediate noise injection are more robust against adversarial attacks compared to models trained with unlabeled data when the Jacobian regularization is utilized. This suggests that noise injection at intermediate layers in the same way the unlabeled data was generated trains more adversarially robust models when using Jacobian regularization term.

\subsection{Prediction Confidence Decreases for Unknown Unknown Classes}
\label{sec:uuc}
Open Set Recognition (OSR) describes the scenario where a DL model is trained with data points from the training dataset known as known known data points (KKCs) similar to normal training. However, during testing, new classes which are not seen during training also known as unknown unknown data points (UUCs) appear among KKC data points. The model is expected to correctly identify KKC data points and reject UUC data points \cite{geng2020recent}. Many proposed methods train DL models with an additional class label for the UUCs and aim for a decision boundary that tightly binds the support for each class distribution. The work by Shao et al. \cite{shao2020openset} showed that such OSR methods are vulnerable to adversarial attacks and proposed a defense mechanism to detect and remove adversarial noise. Other OSR algorithms do not add a new class label, but instead train models to assign UUC data points to every class with low probability \cite{scheirer2014probability}. Since UUC distributions are hypothesized to be away from the class distributions seen in the training dataset, we evaluated whether our trained models with the proposed training framework are able to provide OSR by decreasing the prediction confidence for UUC data points. \par

\begin{table}[t]
    \centering
    \caption{Table compares prediction confidence of DL models on UUC data points taken from Imagenette dataset. The DL models are trained on MNIST and CIFAR10 datasets using our training framework and without regularization for five different seeds.}
    \includegraphics[clip=true, trim=2cm 7cm 12cm 0cm,width=0.9\textwidth]{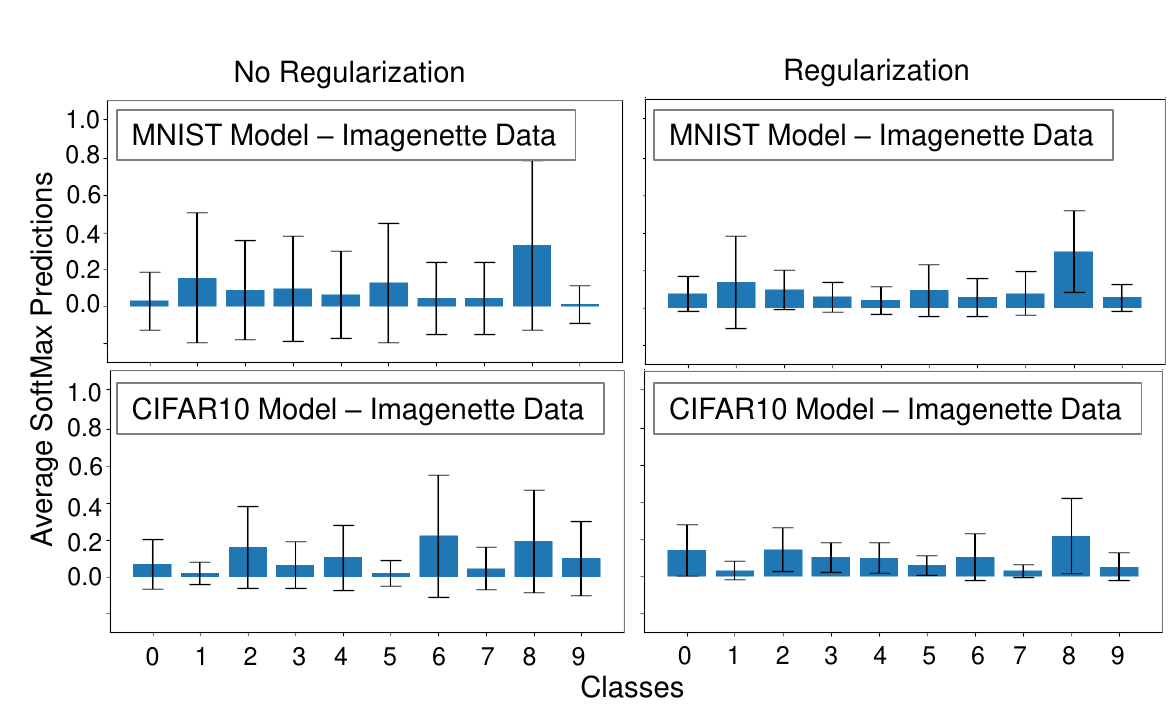}
\label{tab:uuc}
\end{table}

We fed the models trained on the MNIST dataset with the test set for Imagenette dataset and measured the mean and standard deviation of the predictions for each class. We performed the same experiment for the models trained on the CIFAR10 dataset. The results were compared with the mean and standard deviation of predictions for models trained using normal training. Table~\ref{tab:uuc} shows the average class prediction and the standard deviation of predictions for each class. As shown, the standard deviation of predictions for each class decreases and the predictions are spread more evenly among the classes for models trained using regularization and unlabeled data. In other words, the prediction confidence of models on UUC data points is consistently lower for models trained with regularization compared to model trained without regularization implying that the SoftMax score surface at the location of UUC distributions in the high dimensional input data space is flattened. 

\section{Discussion and Conclusion}
In this paper, we hypothesized that data sparsity and large numbers of redundant parameters in DL models allow different DL models to learn decision boundaries with different appearances in the space outside the support of the data distribution. Such models can have equally high prediction accuracy on clean datasets, however, their robustness against adversarial attacks can be very different. We also hypothesized that a model with a smooth SoftMax score surface whose decision boundary stays far from each class distribution is ideal in terms of adversarial robustness. Following these hypotheses, we automatically generated an unlabeled dataset to cover the space outside the support of the data distribution using Gaussian noise. We then used a regularization term to train DL models with smooth SoftMax score surfaces using both the original and unlabeled dataset. Using the generated unlabeled data, we were able to train DL models to have smooth SoftMax score surfaces outside the support of the input data distribution. We empirically showed that the proposed training framework indeed increases the distance from data points to the decision boundary for models trained on MNIST, CIFAR10 and Imagenette datasets. We also showed that the proposed training framework trains models with SoftMax score surfaces similar to the hypothesized ideal SoftMax score surface for a 2D Points dataset. \par
Although we believe the Jacobian regularizer is a strong regularization term which can train DL models with smooth SoftMax score surfaces, we deployed a more naive regularization term with shorter runtime for our rigorous experiments. However, in multiple occasions we deployed the Jacobian regularizer and showed that the latter generally performs better than the former.\par
We also showed that using an unlabeled dataset generated by Gaussian noise while considering the spread of the labeled data distribution in the high dimensional input data space is almost as effective as unlabeled datasets sourced from expensive and manually collected datasets or generated using complex synthesis algorithms. Finally, we demonstrated that models trained with our framework also show consistently less confidence in classifying samples from unknown classes as expected in the open set recognition domain. \par
This paper empirically tests our hypothesis about data sparsity, DL models and their decision boundaries. Our future objective is to theoretically prove that Jacobian regularization and our unlabeled data indeed train models with smooth SoftMax score surface with decision boundaries far from each class distribution.

{\small
\bibliographystyle{ieee}
}

\section*{A - Effect of the Proposed Regularizer on SoftMax Score Surface for the 2D Points Dataset}
In this section, we provide training parameters for the 2D Points set experiment. We also provide more examples of the decision boundaries learned by our model using normal training, Jacobian Regularization \cite{Jakubovitz2018improving} and our proposed training framework.\par
\begin{figure}[ht]
    \centering
    \includegraphics[clip=true, trim=0cm 1cm 10cm 5cm,width=0.8\textwidth]{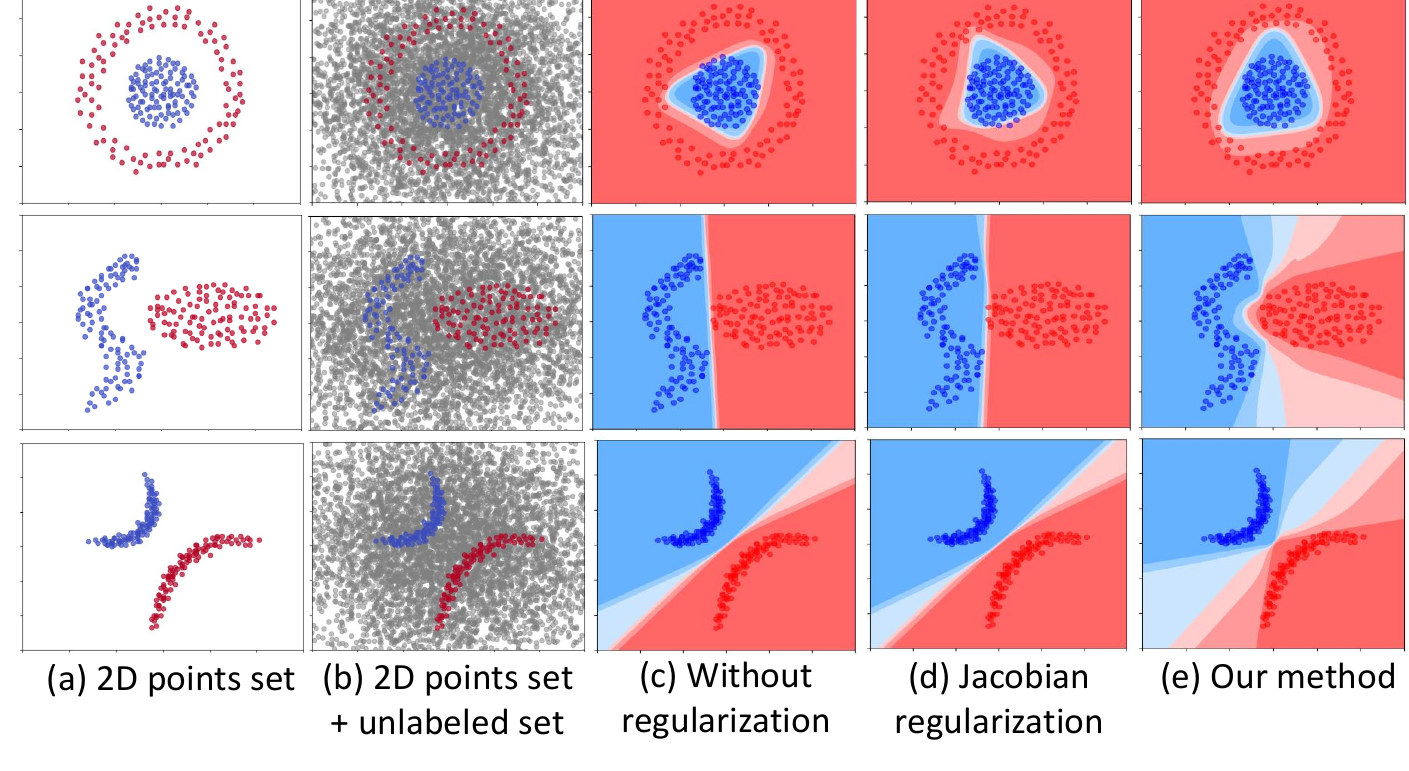}
    \caption{Figure shows more examples of 2D Points sets and the decision boundary learned by models trained with normal training, using Jacobian Regularizer \cite{Jakubovitz2018improving} without unlabeled data and the proposed training framework with unlabeled data. The reader is referred to the online version of this paper for the color representation of this figure. }
    \label{fig:2DExp}
\end{figure}

We trained a neural network with four fully connected layers. The number of neurons in each layer was 2, 8, 8, and 2 respectively. The TanH activation function was used in all experiments on the 2D Points set. Stochastic Gradient Descent with a learning rate of $1 \times 10^{-3}$, momentum of 0.9, and weight decay of $5 \times 10^{-4}$ was used for optimization. The $\lambda$ value was set to 1000 for both the proposed training framework and for Jacobian Regularization \cite{Jakubovitz2018improving} in Fig. 3 of the main paper. Fig. \ref{fig:2DExp} presents more examples of the decision boundaries that models learn when trained with normal training, Jacobian Regularization \cite{Jakubovitz2018improving} with $\lambda=10$ without unlabeled data, and the proposed training framework with $\lambda=1000$ and unlabeled data. 
As shown by these examples, the proposed training framework is able to spread the SoftMax surface more smoothly than the Jacobian regularizer \cite{Jakubovitz2018improving} in the space around the two class distributions without labeled data points. These examples show the advantage of using unlabeled data by our proposed training framework in learning a smooth SoftMax surface in the space around the class distributions. 

\section*{B - Experiment Design for Section 4.2: Numerical Results on the Distance of Decision boundary from Data Distribution}
\label{sec:hypers1}
In this section, we provide the set of hyperparameters used in experiments for each image dataset.
\subsection*{MNIST Dataset}
For experiments on MNIST dataset, LeNet \cite{lecun1998gradient} network architecture was used with Sigmoid activation function and a learning rate of $10^{-3}$. Adam optimizer was used for training with no weight decay. The mean ($\mu$) values for both Gaussian distributions used to generate unlabeled and neighbor data points were set to zero and the standard deviations ($\sigma$) were set to $0.126$, and 10-times smaller $\sigma$ of $0.0126$, respectively. The model was trained using our proposed loss function. Different values of $\lambda$ were tested and the model's loss curve was checked for each $\lambda$ value. The highest $\lambda$ value that did not decrease the model's validation accuracy significantly was chosen. This value was $100k$ for the MNIST dataset. Models were trained with $\lambda=100k$, number of unlabeled data points per labeled data point $N_u=1$ and number of neighbors per data point $N_b=1$. The models trained using these hyperparameters were used to measure  $\mathbf{\hat{\rho}_{adv}}$ values and test accuracy in Table 1 of the main paper. 

\subsection*{CIFAR10 Dataset}
The models were trained on the CIFAR10 dataset using augmentation methods including random crops, horizontal flipping, and perspective change. The ResNet 9 \cite{he2016deep} network architecture was used with a CELU activation function for all layers. The model was trained with Adam optimizer with a learning rate of $10^{-4}$, no weight decay and the standard deviations for generating unlabeled data points and neighbor data points were set to $0.238$ and $0.0238$, respectively. The lambda value was chosen similarly and was set to $\lambda = 10k$ for the CIFAR10 dataset. Using $\lambda=10k$, the model was trained using the proposed loss function and the generated unlabeled dataset. Models were trained with $\lambda=10k$, number of unlabeled data points per labeled data point $N_u=1$ and number of neighbors per data point $N_b=1$. The models trained using these hyperparameters were used to measure  $\mathbf{\hat{\rho}_{adv}}$ values and test accuracies in Table 2 of the main paper.

\subsection*{Imagenette Dataset}
Imagenette is a subset of the well-known ImageNet ILSVRC 2012 dataset \cite{russakovsky2015imagenet} with 10 easily classified classes: tench, English springer, cassette player, chain saw, church, French horn, garbage truck, gas pump, golf ball, parachute. For each of the 10 classes, Imagenette has around 1000 and 400 data points in the training and validation datasets respectively. The XResNet 18 \cite{he2019bag} network architecture was used to train a model to classify the Imagenette dataset with the MISH activation function. The Adam optimizer was used for training with a learning rate of $1 \times 10^{-4}$ and no weight decay. The standard deviations for the Gaussian distributions for generating unlabeled and neighbor data points were set to $0.139$ and $0.0139$ respectively. We used $\lambda = 10k$ for regularization experiments. The number of unlabeled data points per labeled data point $N_u$ and number of neighbors per labeled data point $N_b$ were set to one. The trained models with these hyperparameters were used to measure  $\mathbf{\hat{\rho}_{adv}}$ values and test accuracies in Table 3.\par
\begin{figure}
    \centering
     \includegraphics[clip=true, trim=4cm 4cm 4cm 0cm,width=\textwidth]{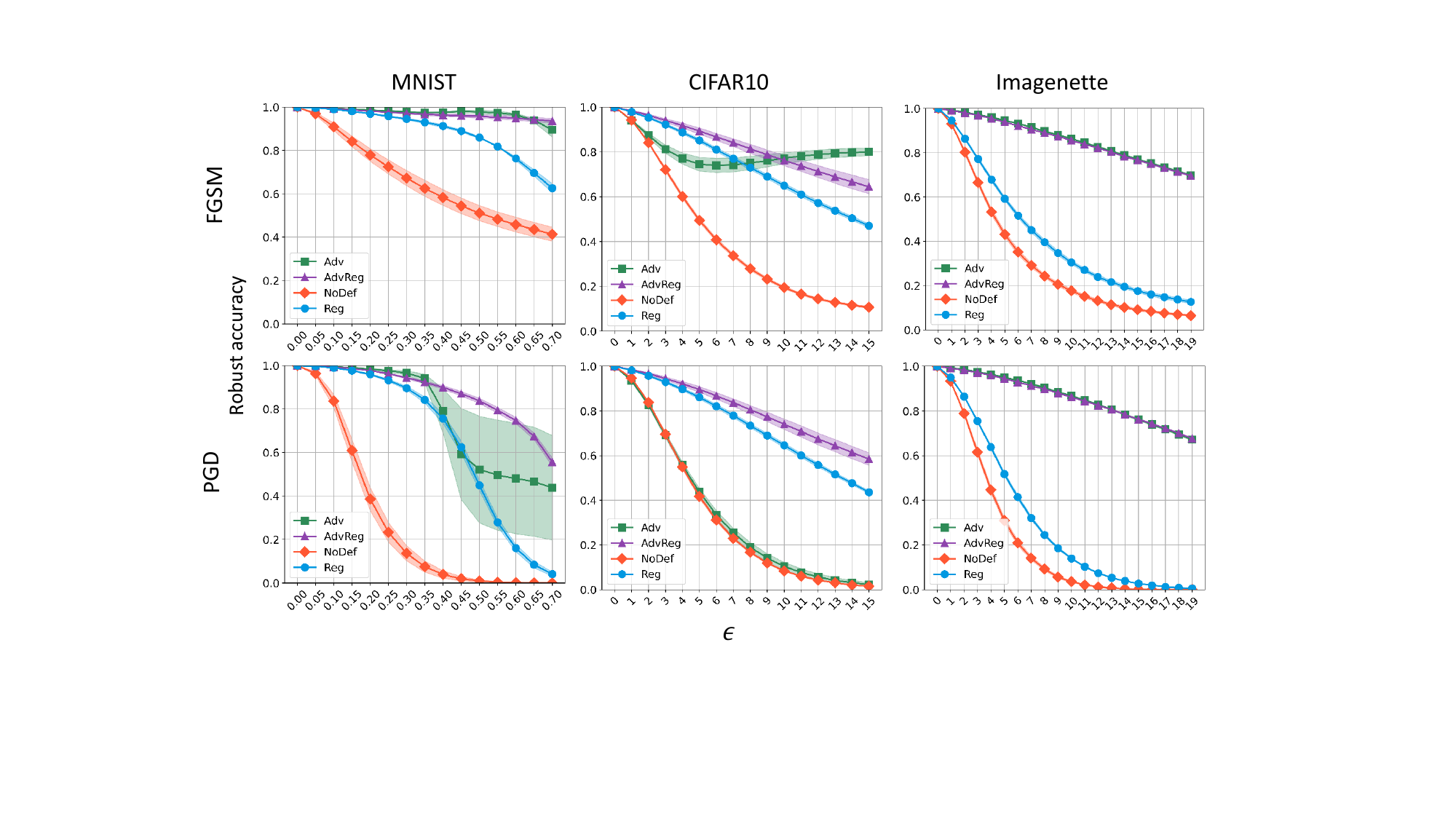}
    \caption{Figure shows the robust accuracy of models trained on the MNIST, CIFAR10 and Imagenette datasets against FGSM \cite{goodfellow2014explaining} and PGD \cite{madry2017towards} attacks with different epsilon perturbation values ($\epsilon$). Robust accuracy of the models against DeepFool \cite{moosavi2016deepfool} ($l^2$ and $l^{\infty}$) are not shown as the robust accuracy stayed almost constant for all models. The reader is referred to the online version of this paper for the color representation of this figure.}
    \label{fig:finalattacks}
\end{figure}

We also show the robust accuracy of these models against FGSM \cite{goodfellow2014explaining}, and PGD \cite{madry2017towards} attacks in Fig. \ref{fig:finalattacks} with different epsilon perturbation values ($\epsilon$). Robust accuracy of the models against DeepFool \cite{moosavi2016deepfool} ($l^2$ and $l^{\infty}$) are not shown here because for all models the robust accuracy stayed almost constant. 

\section*{C - Experiment Design for Section 4.4: Gaussian Noise is as Effective as Costly and More Complex Unlabeled Datasets}
In this section, we provide more details on the hyperparameters for models used in Fig. 7 of the main paper.\par
\subsection*{MNIST Dataset}
In Fig. 7(a) (Left) of the main paper, the effect of different numbers of unlabeled data points per labeled data point ($N_u$) on robust accuracy of the model was tested for MNIST dataset. Models were trained with $N_u$ = 0, 5, and 10 with fixed parameters $\lambda=100k$, number of labeled data points per class $N_l=64$, and number of neighbors per data point $N_b=100$. Five models were trained for each set of parameters, all with similar validation accuracies. The robust accuracy of the model was tested for different epsilon perturbation values ($\epsilon$) using FGSM \cite{goodfellow2014explaining}, PGD \cite{madry2017towards}, and DeepFool ($l^{\infty}$) \cite{moosavi2016deepfool} attacks.\par

In Fig. 7(b) (Left) of the main paper, the effect of regularization was tested on the robust accuracy of models trained with different numbers of labeled data points for the MNIST dataset. Models were trained with ($\lambda=100k$) and without ($\lambda=0$) regularization for different numbers of labeled data points per class ($N_l = 8, 64, 256$), with a fixed number of unlabeled data points ($N_u=10$) and neighbors ($N_b=2000$) per data point. Five different models were trained similarly per set of parameters. All trained models had similar validation accuracies. \par

\subsection*{CIFAR10 Dataset}
In Fig. 7(a) (Right) of the main paper, the effect of different numbers of unlabeled data points on the robust accuracy of models trained on CIFAR10 dataset was evaluated. Models were trained with different numbers of unlabeled data points per labeled data point $N_u = 1, 5, 10$ with $\lambda=10k$, with a fixed number of labeled data points per class $N_l=128$ and neighbors per data point $N_b=1$ for the CIFAR10 dataset. Five different models were trained similarly for every set of parameters. All trained models had similar validation accuracies. \par
In Fig. 7(b) (Right) of the main paper, the effect of regularization was evaluated on the robust accuracy of models trained with different numbers of labeled data points for CIFAR10 dataset. Models were trained with ($\lambda=10k$) and without ($\lambda=0$) regularization, and different numbers of labeled data points per class ($N_l = 128, 256, 512$). The number of unlabeled data points per labeled data point, $N_u$, and neighbors per data point, $N_b$, were fixed to 1 in these experiments. Five different models were trained similarly per set of parameters. All trained models had similar validation accuracies.\par

\section*{D - Experiment Design for Section 4.6: The Effect of the Network Layer where Gaussian Noise is Applied}
We applied intermediate noise injection on the output of the last convolution layer of the model as follows. The labeled image $x_i \in X$, was passed into the model and the intermediate output $x_{i,mid}$ was intercepted. Noise $\Delta_j, j=\{1,\cdots,N_u\}$ were sampled from an isotropic Gaussian distribution with zero mean and co-variance matrix of $\sigma_{u,mid}^2I$ and were added to $x_{i,mid}$ to generate its associated unlabeled data points $\{u_{i,j,mid} = x_{i,mid} + \Delta_j, j=1,\cdots,N_u\}$. In a similar way, for each labeled or unlabeled intermediate output $\psi_{k,mid}$, noise $\delta_j, j=\{1,\cdots,N_b\}$ from an isotropic Gaussian distribution with zero mean and c-variance matrix of $\sigma_{b,mid}=\sigma_{u,mid}/10$ were sampled and added to $\psi_{k,mid}$ to generate the associated neighbor points, $\{\psi_{k,j,mid} = \psi_{k,mid}+\delta_j, j=1,\cdots,N_b\}$, for $\psi_{k,mid}$. The intermediate output $\psi_{k,mid}$ and it's neighbors $\psi_{k,j,mid}$ were then passed through the remainder of the model. The values for $\sigma_{u,mid}$ and $\sigma_{b,mid}$ are analogous to the $\sigma_u$ and $\sigma_b$. These values were calculated in a similar way, with the only difference that the mean pairwise Euclidean distances were calculated over the intermediate model outputs $\psi_{k,mid}$ instead of the labeled images $\psi_k$. The hyperparameters used were the same as those outlined in Appendix B with the exception of the new standard deviations $\sigma_{u,mid}$ and $\sigma_{b,mid}$. 

\begin{table}[t]   
\centering
\caption{Column $\mathbf{\hat{\rho}_{adv}}$ shows test accuracy and measured $\mathbf{\hat{\rho}_{adv}}$ values using the DeepFool attack \cite{moosavi2016deepfool} for the models trained on MNIST, CIFAR10, and Imagenette datasets using our training framework. The standard deviations used to generate unlabeled and neighbor data points, $\sigma_u$ and $\sigma_b$, were varied from 0.1x to 3x the original $\sigma_u$ and $\sigma_b$ from the main paper (shown in bold).}
\resizebox{0.82\columnwidth}{!}{%
\begin{tabular}{ccccccc}
 & \multicolumn{2}{c}{MNIST} & \multicolumn{2}{c}{CIFAR10} & \multicolumn{2}{c}{Imagenette} \\ \hline
\multirow{2}{*}{Defence method} & $\hat{\rho}_{adv} $ & Test & $\hat{\rho}_{adv}$ & Test  & $\hat{\rho}_{adv}$ & Test \\ 
& & accuracy & &  accuracy  &  &  accuracy\\
&$(\times 10^{-2})$ &(\%) & $(\times 10^{-2})$ &(\%) & $(\times 10^{-2})$ & (\%) \\ \hline
No defence & 20.30 & 98.29 & 1.23 & 91.17 & 0.941 & 87.18 \\
Adversarial training & 61.84 & 98.66 & 1.27 & 89.51 & 4.82 & 83.06 \\ \hline
 & \multicolumn{2}{c}{\begin{tabular}[c]{@{}c@{}}$\sigma_u = 1.26e\text{-}3$,\\ $\sigma_b = 1.26e\text{-}4$ (0.1x)\end{tabular}} & \multicolumn{2}{c}{\begin{tabular}[c]{@{}c@{}}$\sigma_u = 2.38e\text{-}2$,\\ $\sigma_b = 2.38e\text{-}3$ (0.1x)\end{tabular}} & \multicolumn{2}{c}{\begin{tabular}[c]{@{}c@{}}$\sigma_u = 1.39e\text{-}2$,\\ $\sigma_b = 1.39e\text{-}3$ (0.1x)\end{tabular}} \\ \hline
Reg & 44.96 & 98.65 & 3.20 & 81.20 & 1.73 & 86.06 \\
\cline{0-0}
Reg \& Adversarial & \multirow{2}{*}{47.91} & \multirow{2}{*}{98.58} & \multirow{2}{*}{4.63}
& \multirow{2}{*}{83.99} & \multirow{2}{*}{5.01} & \multirow{2}{*}{80.82} \\
 training&  &  &  &  &  &  \\ \hline
 & \multicolumn{2}{c}{\begin{tabular}[c]{@{}c@{}}$\sigma_u = 6.30e\text{-}3$,\\ $\sigma_b = 6.30e\text{-}4$ (0.5x)\end{tabular}} & \multicolumn{2}{c}{\begin{tabular}[c]{@{}c@{}}$\sigma_u = 1.19e\text{-}1$,\\ $\sigma_b = 1.19e\text{-}2$ (0.5x)\end{tabular}} & \multicolumn{2}{c}{\begin{tabular}[c]{@{}c@{}}$\sigma_u = 6.94e\text{-}2$,\\ $\sigma_b = 6.94e\text{-}3$ (0.5x)\end{tabular}} \\ \hline
Reg & 44.49 & 98.67 & 3.08 & 80.80 & 2.46 & 83.44 \\
\cline{0-0}
Reg \& & Adversarial\multirow{2}{*}{50.53} & \multirow{2}{*}{98.76} & \multirow{2}{*}{4.70} & \multirow{2}{*}{84.97} & \multirow{2}{*}{5.31} & \multirow{2}{*}{81.94} \\
 training &  &  &  &  &  &  \\ \hline
 & \multicolumn{2}{c}{\textbf{\begin{tabular}[c]{@{}c@{}}$\mathbf{\sigma_u = 1.26e\text{-}1}$,\\ $\mathbf{\sigma_b=1.26e\text{-}2}$ (1x)\end{tabular}}} & \multicolumn{2}{c}{\textbf{\begin{tabular}[c]{@{}c@{}}$\mathbf{\sigma_u = 2.38e\text{-}1}$,\\ $\mathbf{\sigma_b = 2.38e\text{-}2}$ (1x)\end{tabular}}} & \multicolumn{2}{c}{\textbf{\begin{tabular}[c]{@{}c@{}}$\mathbf{\sigma_u = 1.39e\text{-}1}$,\\ $\mathbf{\sigma_b = 1.39e\text{-}2}$ (1x)\end{tabular}}} \\ \hline
Reg & \textbf{46.29} & \textbf{98.69} & \textbf{3.29} & \textbf{83.96} & \textbf{2.21} & \textbf{84.92} \\
\cline{0-0}
Reg \& Adversarial& \multirow{2}{*}{\textbf{51.20}} & \multirow{2}{*}{\textbf{98.80}} & \multirow{2}{*}{\textbf{4.45}} & \multirow{2}{*}{\textbf{83.50}} & \multirow{2}{*}{\textbf{4.96}} & \multirow{2}{*}{\textbf{82.44}} \\
 training &  &  &  &  &  &  \\ \hline
 & \multicolumn{2}{c}{\begin{tabular}[c]{@{}c@{}}$\sigma_u = 1.89e\text{-}1$,\\ $\sigma_b = 1.89e\text{-}2$ (1.5x)\end{tabular}} & \multicolumn{2}{c}{\begin{tabular}[c]{@{}c@{}}$\sigma_u = 3.57e\text{-}1$,\\ $\sigma_b = 3.57e\text{-}2$ (1.5x)\end{tabular}} & \multicolumn{2}{c}{\begin{tabular}[c]{@{}c@{}}$\sigma_u = 2.08e\text{-}1$,\\ $\sigma_b = 2.08e\text{-}2$ (1.5x)\end{tabular}} \\ \hline
Reg & 46.63 & 98.63 & 2.94 & 86.16 & 1.85 & 86.01 \\
\cline{0-0}
Reg \& Adversarial& \multirow{2}{*}{60.22} & \multirow{2}{*}{98.21} & \multirow{2}{*}{4.32} & \multirow{2}{*}{86.25} & \multirow{2}{*}{5.53} & \multirow{2}{*}{82.04} \\
 training &  &  &  &  &  &  \\ \hline
 & \multicolumn{2}{c}{\begin{tabular}[c]{@{}c@{}}$\sigma_u = 2.52e\text{-}1$,\\ $\sigma_b = 2.52e\text{-}2$ (2x)\end{tabular}} & \multicolumn{2}{c}{\begin{tabular}[c]{@{}c@{}}$\sigma_u = 4.76e\text{-}1$,\\ $\sigma_b = 4.76e\text{-}2$ (2x)\end{tabular}} & \multicolumn{2}{c}{\begin{tabular}[c]{@{}c@{}}$\sigma_u = 2.77e\text{-}1$,\\ $\sigma_b = 2.77e\text{-}2$ (2x)\end{tabular}} \\ \hline
Reg & 47.22 & 98.52 & 3.08 & 86.60 & 1.41 & 84.46 \\
\cline{0-0}
Reg \& Adversarial& \multirow{2}{*}{50.85} & \multirow{2}{*}{98.71} & \multirow{2}{*}{4.75} & \multirow{2}{*}{86.25} & \multirow{2}{*}{5.52} & \multirow{2}{*}{82.96} \\
 training &  &  &  &  &  &  \\ \hline
 & \multicolumn{2}{c}{\begin{tabular}[c]{@{}c@{}}$\sigma_u = 3.78e\text{-}1$,\\ $\sigma_b = 3.78e\text{-}2$ (3x)\end{tabular}} & \multicolumn{2}{c}{\begin{tabular}[c]{@{}c@{}}$\sigma_u = 7.13e\text{-}1$,\\ $\sigma_b = 7.13e\text{-}2$ (3x)\end{tabular}} & \multicolumn{2}{c}{\begin{tabular}[c]{@{}c@{}}$\sigma_u = 4.16e\text{-}1$,\\ $\sigma_b = 4.16e\text{-}2$ (3x)\end{tabular}} \\ \hline
Reg & 45.68 & 98.65 & 3.04 & 86.16 & 1.35 & 86.32 \\
\cline{0-0}
Reg \& Adversarial& \multirow{2}{*}{60.40} & \multirow{2}{*}{98.19} & \multirow{2}{*}{4.64} & \multirow{2}{*}{86.39} & \multirow{2}{*}{5.10} & \multirow{2}{*}{82.09} \\
 training &  &  &  &  &  &  \\ \cline{1-7}
\end{tabular}}
\label{tab:combined}
\end{table}
\begin{figure}[t]
    \centering
    \includegraphics[clip=true, trim=2.8cm 0.1cm 2.2cm 0cm,width=\textwidth]{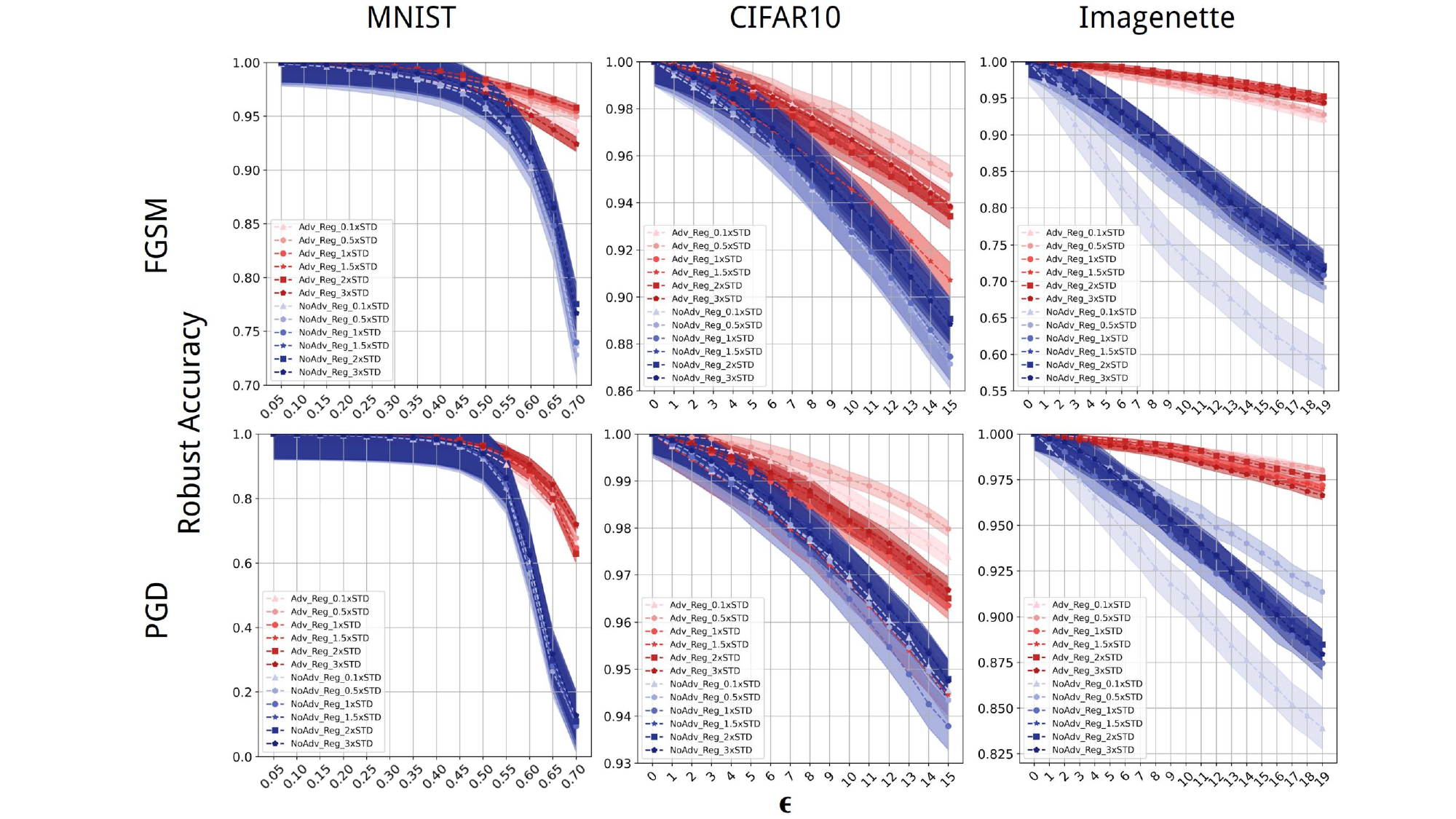}
    \caption{Figure shows the robust accuracy of models trained using the naive regularization and unlabeled data (NoAdv\_Reg) versus models trained using the naive regularization, unlabeled data and  adversarial training (Adv\_Reg). The standard deviations $\sigma_u$ and $\sigma_b$ used to generate unlabeled and neighbor data points respectively were varied from 0.1x to 3x their original values used in the main paper. The robust accuracies were measured using FGSM \cite{goodfellow2014explaining} and PGD \cite{madry2017towards} attacks with different epsilon perturbation values ($\epsilon$). The reader is referred to the online version of this paper for the color representation of this figure.}
    \label{fig:varyingstds}
\end{figure}
\section*{E - The Effect of the Choice of Gaussian Noise on Robustness of Models}
To analyse the impact of the choice of Gaussian distribution on the robust accuracy of models trained using our framework, models were trained with noise sampled from Gaussian distributions with different standard deviations. For reference, $\sigma_u$ and $\sigma_b$ denote the standard deviations to generate unlabeled and neighbor data points respectively.\par
For all three datasets, we performed five additional experiments with $\sigma_u$ and $\sigma_b$ being 0.1x, 0.5x, 1.5x, 2x and 3x the original $\sigma_u$ and $\sigma_b$ used in the main paper. The hyperparameters chosen were the same as those in Appendix B, and three models were trained per experiment. Table \ref{tab:combined}  shows the $\hat{\rho}_{adv}$ and test accuracy for these models. The results for the models that were presented in the main paper are shown in bold. Figure \ref{fig:varyingstds} shows the robust accuracy for these models.\par
The results from Table \ref{tab:combined} show that when only the native regularizer and unlabeled dataset is used in training, the choice of $\sigma_u$ and $\sigma_b$ is important as it does moderate adversarial robustness. Should one choose values of $\sigma_u$ and $\sigma_b$ that are too high or low, adversarial robustness would decrease. As seen from all the rows labeled as "Reg", for MNIST, the multiple of $\sigma_u$ and $\sigma_b$ that gives the highest $\hat{\rho}_{adv}$ is 2x. For CIFAR10, it is 1x. Lastly, for Imagenette, it is 0.5x. We believe that robustness drops when the standard deviation is too high, because the generated unlabeled data points are too far away from the original class distributions. Consequently, there are insufficient unlabeled data points around the class distributions for regularization of the decision surfaces in these areas to take place. Additionally, the generated neighbors are situated too far away to fulfill their original purpose of being in close proximity with the training data points to make the decision surface around each data point smooth. \par

Likewise, if the standard deviation is too low, the unlabeled data points will likely be too close to the original data points, therefore they cannot support enough of the high dimensional input data space. Interestingly, however, according to Figure \ref{fig:varyingstds}, the robust accuracy stayed roughly the same for all the models trained with only naive regularization and unlabeled dataset. This means that whilst the robustness of our models to the DeepFool \cite{moosavi2016deepfool} attack and hence $\hat{\rho}_{adv}$ values are sensitive to the choice of $\sigma_u$ and $\sigma_b$, robustness of our models to FGSM \cite{goodfellow2014explaining} and PGD \cite{madry2017towards} remained largely independent of $\sigma_u$ and $\sigma_b$. Why this is the case remains to be seen. \par
On the other hand, when the naive regularization and unlabeled data was combined with adversarial training, increasing or decreasing $\sigma_u$ and $\sigma_b$ had no discernible effect on the $\hat{\rho}_{adv}$ of models from the 3 datasets. This is evident in Table \ref{tab:combined} and \ref{fig:varyingstds} where the variation in $\hat{\rho}_{adv}$ values and the robust accuracy curves followed no clear trend. This could be because the adversarial examples generated during adversarial training were already very scattered in the input space. Therefore, further variations of $\sigma_u$ or $\sigma_b$ did not impact adversarial robustness significantly. \par

\end{document}